\documentclass{article}

\usepackage[main, final]{neurips_2026}

\usepackage[utf8]{inputenc}
\usepackage[T1]{fontenc}
\usepackage{hyperref}
\usepackage{graphicx}

\usepackage{url}
\usepackage{booktabs}
\usepackage{amsfonts}
\usepackage{amsmath,amssymb,amsthm,mathtools,bm}
\usepackage{aliascnt}
\usepackage{nicefrac}
\usepackage{microtype}
\usepackage{xcolor}
\usepackage{enumitem}
\usepackage{algorithm}
\usepackage{algpseudocode}
\usepackage[noabbrev]{cleveref}
\usepackage{tikz}
\usetikzlibrary{arrows.meta,calc,positioning,shapes.geometric,fit,backgrounds}
\crefname{equation}{equation}{equations}
\Crefname{equation}{Equation}{Equations}
\crefname{appendix}{appendix}{appendices}
\Crefname{appendix}{Appendix}{Appendices}

\title{Resolvent Ellipsoid for Minty Set Inclusions}

\author{%
  Ashkan Soleymani \qquad
  Gabriele Farina \qquad
  Patrick Jaillet\\[0.4em]
  {\normalfont Massachusetts Institute of Technology}\\[0.2em]
  {\normalfont\texttt{\{ashkanso,gfarina,jaillet\}@mit.edu}}%
}

\newtheorem{theorem}{Theorem}
\newaliascnt{proposition}{theorem}
\newtheorem{proposition}[proposition]{Proposition}
\aliascntresetthe{proposition}
\newaliascnt{lemma}{theorem}
\newtheorem{lemma}[lemma]{Lemma}
\aliascntresetthe{lemma}
\newaliascnt{corollary}{theorem}
\newtheorem{corollary}[corollary]{Corollary}
\aliascntresetthe{corollary}
\newaliascnt{remark}{theorem}
\newtheorem{remark}[remark]{Remark}
\aliascntresetthe{remark}
\newaliascnt{assumption}{theorem}
\newtheorem{assumption}[assumption]{Assumption}
\aliascntresetthe{assumption}
\newaliascnt{definition}{theorem}
\newtheorem{definition}[definition]{Definition}
\aliascntresetthe{definition}
\newaliascnt{claim}{theorem}

\aliascntresetthe{claim}
\newaliascnt{example}{theorem}

\aliascntresetthe{example}
\crefname{theorem}{theorem}{theorems}
\Crefname{theorem}{Theorem}{Theorems}
\crefname{proposition}{proposition}{propositions}
\Crefname{proposition}{Proposition}{Propositions}
\crefname{lemma}{lemma}{lemmas}
\Crefname{lemma}{Lemma}{Lemmas}
\crefname{corollary}{corollary}{corollaries}
\Crefname{corollary}{Corollary}{Corollaries}
\crefname{remark}{remark}{remarks}
\Crefname{remark}{Remark}{Remarks}
\crefname{assumption}{assumption}{assumptions}
\Crefname{assumption}{Assumption}{Assumptions}
\crefname{definition}{definition}{definitions}
\Crefname{definition}{Definition}{Definitions}
\crefname{claim}{claim}{claims}
\Crefname{claim}{Claim}{Claims}
\crefname{example}{example}{examples}
\Crefname{example}{Example}{Examples}

\colorlet{ellfill}{black!12}
\colorlet{mintyred}{red!75!black}
\colorlet{stepblue}{blue!70!black}
\colorlet{cutred}{red!70!black}
\colorlet{thickblue}{blue!45}
\colorlet{goodgreen}{green!50!black}

\definecolor{myOrange}{RGB}{235, 105, 92}
\definecolor{myMaroon}{RGB}{128, 0, 0}
\definecolor{mygreen}{RGB}{160, 200, 140}

\hypersetup{ 
colorlinks=true, 
linkcolor=myOrange, 
citecolor=myMaroon, filecolor=cyan, 
urlcolor=mygreen
}

\DeclareMathOperator{\gra}{gra}
\DeclareMathOperator{\dist}{dist}
\DeclareMathOperator{\argmin}{argmin}
\DeclareMathOperator{\vol}{vol}

\newcommand{\R}{\mathbb{R}}
\newcommand{\X}{\mathcal{X}}
\newcommand{\eps}{\varepsilon}
\newcommand{\ip}[2]{\left\langle #1, #2\right\rangle}
\newcommand{\norm}[1]{\left\|#1\right\|}
\newcommand{\set}[1]{\left\{#1\right\}}

\newcommand{\Gap}{\mathrm{Gap}}
\newcommand{\Minty}{\mathcal{K}}
\newcommand{\ball}[2]{B_{#2}(#1)}
\newcommand{\simplex}[1]{\Delta_{#1}}

\newtheorem*{repeatthminner}{\repeatthmname}
\newcommand{\repeatthmname}{}

\newenvironment{repeatthm}[1]{%
  \renewcommand{\repeatthmname}{Theorem~\ref{#1}}%
  \begin{repeatthminner}%
}{%
  \end{repeatthminner}%
}

\newtheorem*{repeatlemmainner}{\repeatlemmaname}
\newcommand{\repeatlemmaname}{}

\newenvironment{repeatlemma}[1]{%
  \renewcommand{\repeatlemmaname}{Lemma~\ref{#1}}%
  \begin{repeatlemmainner}%
}{%
  \end{repeatlemmainner}%
}

\newtheorem*{repeatpropinner}{\repeatpropname}
\newcommand{\repeatpropname}{}

\newtheorem*{repeatcorollaryinner}{\repeatcorollaryname}
\newcommand{\repeatcorollaryname}{}

\def\[#1\]{\begin{align*}#1\end{align*}}

\NewDocumentCommand{\numberthis}{om}{%
  \IfNoValueTF{#1}{%
    \refstepcounter{equation}\tag{\theequation}%
  }{%
    \tag{#1}%
  }%
  \label{#2}%
}

\begin{document}

\maketitle

\begin{abstract}
We study Minty set inclusion on a compact convex set $\X \subseteq \R^d$ with a set-valued operator $A : \X \rightrightarrows \R^d$. Under the Minty promise, the goal is to compute $x \in \X$ and $u \in A(x)$ such that $\sup_{y \in \X} \langle u, x - y \rangle \le \eps$. Our approach is geometric and we work in the fixed-scale constrained-resolvent oracle model, where a query at anchor $a$ is assumed to return a resolvent point $x^+$ together with the Yosida vector $g_{\eta}(a) = (a - x^+)/\eta$. We prove that each oracle response yields a sharp dichotomy: either the returned point provides an $\eps$-SVI certificate, or the Yosida vector defines a strict separating normal for the Minty set. Thus, a single black-box resolvent query supplies either a certified candidate solution or a valid geometric cut. Combining this separation principle with the ellipsoid method, we obtain \textsc{Resolvent-Ellipsoid}, an algorithm that computes an $\eps$-SVI in $\mathcal{O}\!\left(\operatorname{poly}(d, \log(1/\eps))\right)$ black-box oracle calls to the fixed-scale constrained resolvent. For monotone inclusions, our algorithm gives, to our knowledge, the first $\mathcal{O}\!\left(\mathrm{poly}(d, \log(1/\eps))\right)$ black-box guarantee in the fixed-scale resolvent model.
\end{abstract}

\section{Introduction}

Black-box resolvent access is a classical oracle model for monotone inclusion problems, dating back to the proximal point method of \citet{rockafellar1976monotone}. In this model, the basic oracle operation is to solve a regularized inclusion (resolvent subproblem), whereas the standard first-order oracle for variational inequalities evaluates a single-valued operator at a given point.
For classical proximal-point methods, the known worst-case convergence rates under monotonicity alone are sublinear. After $k$ resolvent evaluations, the proximal point method has a tight $\mathcal{O}(1/k)$ rate for the squared fixed-point residual \citep{gu2020tight}, while accelerated variants achieve $\mathcal{O}(1/k^2)$ \citep{kim2021accelerated,park2022exact}. These rates translate into $\mathcal{O}(1/\eps)$ and $\mathcal{O}(1/\sqrt{\eps})$ resolvent evaluations, respectively, to obtain an iterate whose squared fixed-point residual is at most $\eps$. Moreover, \citet{park2022exact} establish matching complexity bounds for general nonexpansive fixed-point schemes. Linear convergence, which yields logarithmic dependence on $1/\eps$, is typically established under additional regularity assumptions, such as strong monotonicity, contractivity, metric subregularity, or H\"older-type growth conditions \citep{rockafellar1976monotone,leventhal2009metric,kim2021accelerated,park2022exact}.

For variational inequalities (VIs) with a single-valued operator $F:\X\to\R^d$, the standard first-order oracle returns $F(x)$ at a queried point $x$. For Lipschitz monotone operators, Halpern-type methods achieve near-optimal $\widetilde{\mathcal{O}}(1/\eps)$ complexity in terms of an operator residual \citep{diakonikolas2020halpern}. More recently, \citet{anagnostides2025polynomial} obtained the first $\operatorname{poly}(d,\log(1/\eps))$ point-evaluation algorithm for Lipschitz VIs under the Minty condition. Their algorithm computes an $\eps$-approximate solution of the Stampacchia variational inequality ($\eps$-SVI), namely a point $x\in\X$ satisfying $\sup_{y\in\X}\langle F(x),x-y\rangle\le\eps$. The main geometric challenge is to construct separating cuts with positive margin. A cut through an ellipsoid center $a_t$ with normal $F(a_t)$ may pass through a Minty solution and therefore fail to preserve any positive-radius neighborhood of the Minty set. The extra-gradient ellipsoid method addresses this issue by evaluating $F$ at a corrected point $\tilde a_t$ and using $F(\tilde a_t)$ as the cutting normal.

For structured inclusion problems arising in composite optimization, primal-dual saddle-point systems, and KKT formulations, computation often relies on regularized subproblem solves, such as proximal maps, splitting steps, or conic routines for strongly regularized instances. Can the polylogarithmic dependence on $1/\eps$ achieved for single-valued Minty variational inequalities also be obtained for set-valued Minty inclusions using only fixed-scale resolvent queries?

\begin{figure}[t]
\centering
\resizebox{0.99\linewidth}{!}{%
\begin{tikzpicture}[>=Latex, line cap=round, line join=round, every node/.style={font=\small}]

\node[font=\bfseries\normalsize] at (2.65,4.48) {Naive Center Cut};
\node[font=\itshape\small,align=center] at (2.65,4.04) {cut with $F(a_t)$, margin $0$};
\node[font=\bfseries\normalsize] at (8.20,4.48) {Extra-gradient};
\node[font=\itshape\small,align=center] at (8.20,4.04) {move along $-F(a_t)$, cut with $F(\tilde a_t)$};
\node[font=\bfseries\normalsize] at (13.75,4.48) {Resolvent};
\node[font=\itshape\small,align=center] at (13.75,4.04) {one solve returns $x_t^+$ and $g_t$};

\begin{scope}[shift={(0.12,0.28)}]
  \filldraw[fill=ellfill,draw=black,line width=0.9pt] (2.65,1.86) ellipse (2.18 and 1.32);
  \node at (0.83,3.10) {$\mathcal{E}_t\subseteq\mathcal{X}$};

  \coordinate (xstar1) at (1.26,1.26);
  \coordinate (a1) at (3.34,1.58);
  \coordinate (Fa1tip) at (3.48,0.66);

  \node[diamond,fill=mintyred,draw=mintyred,inner sep=1.6pt] (mx1) at (xstar1) {};
  \node[mintyred,above left=-1pt and -1pt] at (mx1) {$x^\star$};

  \fill (a1) circle (1.35pt);
  \node[above right=-1pt and -1pt] at (a1) {$a_t$};

  \draw[black!70,dashed,line width=0.95pt] (0.55,1.18) -- (4.85,1.78);
  \draw[->,black!85,line width=1.0pt] (a1) -- (Fa1tip);
  \node[black!85,right=1pt] at ($(a1)+(0.10,-1.05)$) {$F(a_t)$};
\end{scope}

\begin{scope}[shift={(5.58,0.28)}]
  \filldraw[fill=ellfill,draw=black,line width=0.9pt] (2.65,1.86) ellipse (2.18 and 1.32);
  \node at (0.83,3.10) {$\mathcal{E}_t\subseteq\mathcal{X}$};

  \coordinate (xstar2) at (1.26,1.26);
  \coordinate (a2) at (3.34,1.58);
  \coordinate (Fa2tip) at (3.48,0.66);

  \coordinate (atilde2) at (3.25,2.18);

  \coordinate (FtildeVec) at (0.92,-0.50);
  \coordinate (Ftilde2tip) at ($(atilde2)+(FtildeVec)$);
  \coordinate (FtildeFromA) at ($(a2)+(FtildeVec)$);

  \node[diamond,fill=mintyred,draw=mintyred,inner sep=1.6pt] (mx2) at (xstar2) {};
  \node[mintyred,above left=-1pt and -1pt] at (mx2) {$x^\star$};

  \fill (a2) circle (1.35pt);
  \node[above right=-1pt and -1pt] at (a2) {$a_t$};

  \fill[stepblue] (atilde2) circle (1.35pt);
  \node[stepblue,above left=1pt and -1pt] at (atilde2) {$\tilde a_t$};

  \draw[->,black!55,line width=0.85pt] (a2) -- (atilde2);
  \node[black!55,left=1pt] at ($(a2)!0.52!(atilde2)+(-0.03,0.01)$) {$-\eta F(a_t)$};

  \draw[->,stepblue,line width=1.0pt] (atilde2) -- (Ftilde2tip);
  \draw[->,stepblue,line width=1.0pt] (a2) -- (FtildeFromA);

  \node[stepblue,right=-1pt] at ($(atilde2)+(0.85,-0.54)$) {$F(\tilde a_t)$};
  \node[stepblue,right=-1pt] at ($(a2)+(0.85,-0.66)$) {$F(\tilde a_t)$};

  \draw[stepblue,line width=0.95pt] ($(a2)+(-0.84,-1.50)$) -- ($(a2)+(0.80,1.45)$);
  \draw[stepblue,dashed,line width=0.90pt] ($(xstar2)+(-0.58,-1.05)$) -- ($(xstar2)+(1.14,2.03)$);

  \draw[<->,stepblue,line width=0.95pt] (1.00,0.92) -- (2.47,0.10);
  \node[stepblue,below=2pt] at (1.74,0.51) {$\geq \gamma_t$};
\end{scope}

\begin{scope}[shift={(11.04,0.28)}]
  \filldraw[fill=ellfill,draw=black,line width=0.9pt] (2.65,1.86) ellipse (2.18 and 1.32);
  \node at (0.83,3.10) {$\mathcal{E}_t\subseteq\mathcal{X}$};

  \coordinate (xstar3) at (1.27,1.27);
  \coordinate (a3) at (3.36,1.58);
  \coordinate (xp3) at (2.65,1.48);
  \coordinate (g3tip) at (4.55,1.74);

  \node[diamond,fill=mintyred,draw=mintyred,inner sep=1.6pt] (mx3) at (xstar3) {};
  \node[mintyred,above left=-1pt and -1pt] at (mx3) {$x^\star$};

  \fill (a3) circle (1.35pt);
  \node[above right=-1pt and -1pt] at (a3) {$a_t$};

  \fill[goodgreen] (xp3) circle (1.35pt);
  \node[goodgreen,below left=0pt and -1pt] at (xp3) {$x_t^+$};

  \draw[->,goodgreen!70!black,line width=0.90pt] (xp3) -- (a3);
  \node[goodgreen,below=2pt] at (2.99,1.39) {$\eta g_t$};

  \draw[->,goodgreen,line width=1.0pt] (a3) -- (g3tip);
  \node[goodgreen,right=0pt] at (4.36,1.88) {$g_t$};

  \draw[goodgreen,line width=0.95pt] ($(a3)+(-0.18,1.55)$) -- ($(a3)+(0.14,-1.40)$);
  \draw[goodgreen,dashed,line width=0.90pt] ($(xstar3)+(-0.18,1.55)$) -- ($(xstar3)+(0.14,-1.40)$);

  \coordinate (smargL) at (1.34,0.28);
  \coordinate (smargR) at ($(smargL)+(2.09,0.31)$);
  \draw[<->,goodgreen,line width=0.95pt] (smargL) -- (smargR);
  \node[goodgreen,below=2pt] at ($(smargL)!0.5!(smargR)$) {$\geq s_t$};
\end{scope}

\end{tikzpicture}
}
\caption{How extra-gradient and resolvent repair the same zero-margin obstruction. Left: the naive cut through $a_t$ with normal $F(a_t)$ can pass through a Minty solution $x^\star$, so it separates but does not protect any positive-radius neighborhood of the Minty set. Middle: extra-gradient uses the same center direction $F(a_t)$ as the naive panel, but first moves from $a_t$ to $\tilde a_t$ along $-F(a_t)$ and then cuts through $a_t$ with the corrected normal $F(\tilde a_t)$; the dashed parallel support line through $x^\star$ makes the margin explicit, and $\gamma_t$ is the perpendicular offset between that support line and the strict cut. Right: the constrained resolvent returns $x_t^+=J_{\eta(F+N_{\X})}(a_t)$ and the Yosida vector $g_t=(a_t-x_t^+)/\eta$ in one shot; hence the correction segment $x_t^+\to a_t$ and the normal $g_t$ are collinear, the dashed parallel support line through $x^\star$ shows the cut margin $s_t=\|a_t-x_t^+\|=\eta\|g_t\|$. Extra-gradient and resolvent therefore repair the same geometric failure (lack of margin) by explicit versus implicit correction. 
}
\label{fig:eg-vs-resolvent}
\end{figure}
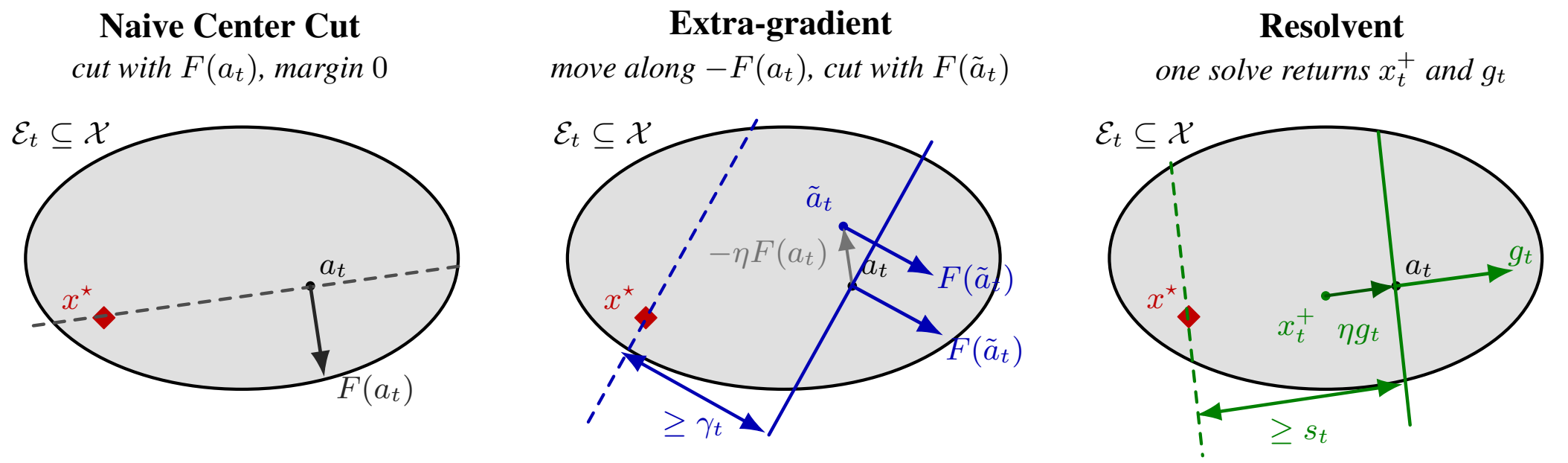

We answer this question affirmatively. We consider a set-valued operator $A:\X\rightrightarrows\R^d$ on a compact convex set $\X$ and study the following Minty inclusion problem: under the promise that the Minty set is nonempty, compute a point $x\in\X$ and a vector $u\in A(x)$ such that
\[
\sup_{y\in\X}\langle u,x-y\rangle\le \eps.
\]
Our main result gives a $\operatorname{poly}(d,\log(1/\eps))$ algorithm in the fixed-scale resolvent model.

The key observation is that a single regularized solve simultaneously produces both an approximate solution and a separating normal. Specifically, if
\[
a=x^+ + \eta(u^+ + n^+), \qquad u^+\in A(x^+),\qquad n^+\in N_{\X}(x^+),
\]
then the Yosida vector
\[
g_{\eta}(a):=u^+ + n^+
\]
provides the candidate point $x^+$ together with the normal that defines the cut. In this sense, one resolvent query automatically performs the correction that extra-gradient methods construct explicitly. \Cref{fig:eg-vs-resolvent} illustrates this shared margin geometry. In the left panel, the naive cut through $a_t$ with normal $F(a_t)$ can pass through a Minty solution $x^\star$, yielding zero margin. In the middle panel, the extra-gradient method moves from $a_t$ to $\tilde a_t$ along the direction $F(a_t)$ and obtains a strict cut whose margin $\gamma_t$ is the perpendicular distance from $x^\star$ to the cut. In the right panel, the resolvent analogue produces a correction segment $x_t^+\to a_t$ that is collinear with the normal $g_t$, and the cut margin is the offset $s_t$. Thus, the resolvent oracle directly provides the information needed to construct a positive-margin cut, without requiring a separate extra-gradient correction.

The geometry is clearest in the unconstrained gradient case \(F=\nabla f\). 
At a center \(a_t\), the naive ellipsoid cut with normal \(\nabla f(a_t)\) may have zero margin: it can separate but still pass through a Minty solution. Extra-gradient repairs this by an \emph{explicit predictor-corrector} step. It first moves \emph{forward} to $
\widetilde a_t = a_t-\eta \nabla f(a_t),$
and then cuts through \(a_t\) using the corrected normal \(\nabla f(\widetilde a_t)\). The Lipschitz assumption is precisely what makes this forward correction controlled and converts separation into separation with positive margin. The resolvent step performs the analogous correction \emph{implicitly}. Instead of evaluating the field at \(a_t\), it computes \(x_t^+\) from the backward equation
\[
x_t^+ = a_t-\eta \nabla f(x_t^+),
\qquad
g_t := \frac{a_t-x_t^+}{\eta}=\nabla f(x_t^+).
\]
Equivalently, when \(f\) is convex, $
x_t^+ \in \arg\min_x
\left\{
f(x)+\frac{1}{2\eta}\|x-a_t\|^2
\right\}
$. Thus the corrected point and the corrected normal are produced in one step: the displacement $ a_t-x_t^+=\eta g_t $ is already collinear with the cutting normal. Consequently, the strict margin is $
s_t=\|a_t-x_t^+\|=\eta\|g_t\|
$.
This is the key distinction. Extra-gradient obtains margin by an \emph{explicit forward correction}, justified by Lipschitz continuity; the resolvent obtains the same geometric effect from an \emph{implicit backward correction} encoded by the proximal solve. This backward formulation is what remains meaningful in the set-valued setting, where there may be no single vector \(F(a_t)\) to evaluate and no Lipschitz continuity assumption to invoke.

We focus on the following set-valued problem.

\begin{definition}[Minty set inclusion] \label{def:svi}
Let $\X\subseteq \R^d$ be compact and convex, and let $A:\X\rightrightarrows \R^d$ have nonempty values.
The goal is to compute, under the promise $\Minty\neq\varnothing$, a point $x\in\X$ and a witness $u\in A(x)$ such that
\[
\sup_{y\in\X}\ip{u}{x-y}\le \eps.
\]
We call such a pair an \emph{$\eps$-Stampacchia solution} or \emph{$\eps$-SVI}.
\end{definition}

When $A(x)=\{F(x)\}$ is single-valued, this is the usual Minty-VI task on $\X$.
The set-valued formulation also captures subdifferentials, composite operators, and KKT maps, which is precisely the setting in which resolvent access is natural~\citep{minty1962monotone,rockafellar1976monotone}. The contribution is therefore complementary to the point-evaluation model of \citet{anagnostides2025polynomial} as we show that fixed-scale regularized solves become a sufficient primitive for Minty-SVI computation in the broader set-valued setting.
\paragraph{Our Contributions.}
We demonstrate on a compact convex domain $\X$, \emph{fixed-scale constrained resolvent access together with linear optimization over $\X$} yields an $\eps$-SVI in $\operatorname{poly}(d,\log(1/\eps))$ time under the Minty promise.
For monotone $A$, the Minty set coincides with the constrained zero set
\[
K \;=\; \{x\in \X : 0\in A(x)+N_{\X}(x)\},
\]
so the framework specializes directly to feasible compact constrained monotone inclusions of the form $0\in A(x)+N_{\X}(x)$. Thus, to the best of our knowledge, the monotone specialization of our framework gives the \emph{first} $\operatorname{poly}(d,\log(1/\eps))$ black-box guarantee in the fixed-scale constrained resolvent model for feasible compact constrained monotone inclusions, without assuming Lipschitz point evaluation, strong monotonicity, cocoercivity, metric subregularity, or any other error-bound or restrictive geometry conditions.

Our other contributions are as follows.
\begin{itemize}[leftmargin=8pt, itemsep=0pt, topsep=0pt, parsep=0pt, partopsep=0pt]
    \item We show that a single constrained resolvent query provides both an acceptance certificate and a strict separator for the Minty set. This is the geometric core of the paper.
    \item We obtain exact and inexact outer theorems with polylogarithmic accuracy dependence. In the nontrivial regime $\eps\le R^2/\eta$, the outer complexity is $
    \mathcal{O}\!\left(d^2\log\frac{R^2}{\eta\eps}\right) $,
    and any $\mathrm{poly}(d,\log(1/\delta))$ routine for approximate constrained resolvents lifts to a $\mathrm{poly}(d,\log(1/\eps))$ algorithm for Minty-SVI. In particular, this brings polylogarithmic-accuracy guarantees into the native resolvent-oracle literature for compact constrained monotone inclusions.
    \item We extend the geometry to a set-valued weak-Minty condition and prove an either-$\eps$-SVI-or-certificate alternative when the Minty condition fails. As a consequence, we broaden the regime of the weak-Minty condition for VIs with polylogarithmic-accuracy tractability (\Cref{rem:weak-minty-comparison}).
    \item We prove a converse on $\rho$-hypomonotone classes and briefly record canonical oracle realizations.
\end{itemize}

The appendix develops the cutting-plane upgrade and exact certificate duality, and presents representative realizations of the resolvent oracle. It also provides a counterexample to a full converse under the Minty assumption alone and establishes a sharp linear convergence rate under a Yosida error bound.

\section{Related Work}

Proximal and resolvent methods have a long history in monotone operator theory, from Moreau's proximal mapping and Rockafellar's proximal point method to splitting and primal--dual schemes such as Douglas--Rachford and Chambolle--Pock \citep{moreau1965proximite,rockafellar1976monotone,lions1979splitting,eckstein1992douglas,chambolle2011first,bauschke2017convex}. These methods rely on regularized subproblem solves, either through the resolvent of the full operator or through proximal evaluations of individual components. Under monotonicity alone, the known worst-case residual convergence rates for classical proximal-point and fixed-point iterations are algebraic, yielding oracle complexity polynomial in $1/\epsilon$ for the corresponding residual criteria \citep{gu2020tight,kim2021accelerated,park2022exact}.

Cutting-plane methods provide another approach to variational inequalities with single-valued operators, using point evaluations of $F$ to construct separating hyperplanes. \citet{luthi1985solution} applied the ellipsoid method to monotone VIs, although the logarithmic-accuracy $\epsilon$-SVI guarantee in that analysis requires strong monotonicity, as noted by \citet{anagnostides2025polynomial}. Subsequent analytic-center methods considered broader classes of VIs. \citet{goffin1997analytic} studied continuous pseudomonotone VIs and obtained nonasymptotic guarantees under the stronger assumption of pseudo-cocoercivity. \citet{nesterov1999homogeneous} developed homogeneous analytic-center cutting-plane methods with efficiency estimates for variational inequalities, feasibility problems, and constrained minimization. \citet{shen2012proximal} considered continuous operators with approximate point evaluations and established convergence of a proximal analytic-center cutting-plane method. The closest result to ours is \citet{anagnostides2025polynomial}, who obtained the first $\operatorname{poly}(d,\log(1/\epsilon))$ algorithm for $\epsilon$-SVIs of single-valued Lipschitz operators under the Minty condition using point-evaluation access. Our work establishes the corresponding polylogarithmic accuracy dependence for set-valued Minty inclusions using fixed-scale resolvent queries.

\section{Notation and Problem Setup}
\label{sec:setup}

\paragraph{Notation.}
We write $\ip{\cdot}{\cdot}$ for the Euclidean inner product and $\norm{\cdot}$ for the Euclidean norm.
For $x\in\R^d$ and $r>0$,
\[
\ball{x}{r}:=\set{z\in\R^d:\norm{z-x}\le r}
\]
denotes the closed Euclidean ball of radius $r$ centered at $x$.
In particular, $\ball{0}{R}$ is the radius-$R$ ball centered at the origin.
For sets $C,D\subseteq\R^d$, we write
$C+D:=\set{c+d:c\in C,\ d\in D}$.
The graph of a set-valued operator is
$\gra A:=\set{(x,u)\in\X\times\R^d:u\in A(x)}$.

We assume throughout that
\begin{equation}
\ball{0}{1}\subseteq \X \subseteq \ball{0}{R},
\qquad R\ge 1.
\label{eq:well-bounded}
\end{equation}
As usual, this normalization can be enforced by a standard affine preprocessing for well-bounded convex sets \citep{groetschel1993geometric}. For $x\in \X$, the normal cone is
\[
N_{\X}(x):=\set{n\in\R^d:\ip{n}{y-x}\le 0\ \text{for all }y\in\X}.
\]

\begin{definition}[Constrained resolvent and Yosida vector]
\label{def:resolvent}
Set $T:=A+N_{\X}$.
For a scale $\eta>0$ and anchor $a\in\R^d$, the constrained resolvent set is
\[
J_{\eta}(a):=(I+\eta T)^{-1}(a)
=
\set{x\in\X:\exists u\in A(x),\ n\in N_{\X}(x)\ \text{s.t.}\ a=x+\eta(u+n)}.
\]
Any choice of $x^+\in J_{\eta}(a)$ together with $u^+\in A(x^+)$ and $n^+\in N_{\X}(x^+)$ satisfying
$a=x^+ + \eta(u^+ + n^+)$
induces the Yosida vector
\[
g_{\eta}(a):=u^+ + n^+ = \frac{a-x^+}{\eta}.
\]
\end{definition}

When $T$ is maximal monotone, $J_\eta$ is the classical single-valued resolvent and $g_\eta$ is the corresponding Yosida approximation \citep{moreau1965proximite,rockafellar1976monotone,bauschke2017convex}. Under the Minty assumption alone, the same formulas apply, but $J_\eta(a)$ may contain multiple points. The oracle selects a solution $x^+\in J_\eta(a)$, which determines the Yosida vector $g_\eta(a)$.

\begin{definition}[Minty set]
\label{def:minty-set}
The Minty set of $A$ on $\X$ is
\[
\Minty:=\set{x\in\X:\ip{v}{y-x}\ge 0\ \text{for all }y\in\X,\ v\in A(y)}.
\]
\end{definition}

The Minty set is always compact and convex because it is an intersection of $\X$ with closed halfspaces indexed by graph points of $A$; the short proof is deferred to \Cref{app:basic-geometry}.

We work in the fixed-scale model. Exact resolvent access at arbitrarily large scales makes the approximation problem essentially one-shot; \Cref{app:one-shot} records this observation.
We therefore study fixed-scale access.

\begin{assumption}[Exact resolvent oracle]
\label{ass:exact-resolvent}
Fix $\eta>0$.
For every rational anchor $a\in\R^d$, an oracle returns a triple
$(x^+,u^+,n^+)$ satisfying
\[
x^+\in \X,\qquad
u^+\in A(x^+),\qquad
n^+\in N_{\X}(x^+),\qquad
a=x^+ + \eta(u^+ + n^+).
\]
We write $x^+=J_{\eta}(a)$ for the returned selection and
\[
g_{\eta}(a):=u^+ + n^+ = \frac{a-x^+}{\eta}.
\]
We also assume access to linear optimization over $\X$.
\end{assumption}

Black-box resolvent oracle access is a standard algorithmic primitive in the study of monotone inclusion problems~\citet{minty1962monotone,rockafellar1976monotone,luque1984asymptotic,qin2010approximating,briceno2011monotone+,boct2014forward,gu2020tight,bui2020warped,kim2021accelerated,bauschke2021generalized,park2022exact,sadiev2024stochastic,boct2026tikhonov}. For the definition of the $\delta$-approximate resolvent oracle, we refer to \Cref{ass:approx-resolvent} in \Cref{app:basic-geometry}.

\section{Resolvent Geometry}
\label{sec:geometry}

Fix $\eta>0$ and an exact resolvent selection as in \Cref{ass:exact-resolvent}.
For an anchor $a\in\R^d$, let
\[
x^+ := J_{\eta}(a),
\qquad
g_{\eta}(a):=\frac{a-x^+}{\eta}.
\]

\begin{definition}[Yosida gap]
\label{def:gap}
For an anchor $a\in\R^d$, define
\[
\Gap_{\eta}(a):=\max_{y\in\X}\ip{g_{\eta}(a)}{a-y}.
\]
\end{definition}

The first lemma is the acceptance test that serves as a Gap certificate.

\begin{lemma}
\label{lem:gap-cert}
If $\Gap_{\eta}(a)\le \eps$, then the resolvent point $x^+=J_{\eta}(a)$ is an $\eps$-SVI point.
\end{lemma}

\begin{proof}
Write $g_{\eta}(a)=u^+ + n^+$ with $u^+\in A(x^+)$ and $n^+\in N_{\X}(x^+)$.
For any $y\in\X$,
\[
\ip{n^+}{y-x^+}\le 0
\qquad\Longrightarrow\qquad
\ip{u^+}{x^+-y}\le \ip{u^+ + n^+}{x^+-y}.
\]
Using $a=x^+ + \eta g_{\eta}(a)$,
\[
\ip{u^+}{x^+-y}
\le
\ip{g_{\eta}(a)}{a-y} - \ip{g_{\eta}(a)}{a-x^+}
=
\ip{g_{\eta}(a)}{a-y}-\eta\norm{g_{\eta}(a)}^2
\le
\Gap_{\eta}(a).
\]
Thus $\ip{u^+}{x^+-y}\le \eps$ for all $y\in\X$.
\end{proof}

The next lemma is the key geometric observation that
the constrained resolvent itself gives a strict separating direction for the Minty set.

\begin{lemma}[Exact strict separation]
\label{lem:exact-separation}
Let $a\in\R^d$ and let $x^+=J_{\eta}(a)$.
For every $x\in\Minty$,
\[
\ip{g_{\eta}(a)}{a-x}
\ge
\eta \norm{g_{\eta}(a)}^2.
\]
Equivalently, if 
$
c(a):=\frac{g_{\eta}(a)}{\norm{g_{\eta}(a)}},
$
then
\[
\ip{c(a)}{x-a}
\le
-\eta\norm{g_{\eta}(a)}.
\]
\end{lemma}

\begin{proof}
Write $g_{\eta}(a)=u^+ + n^+$ with $u^+\in A(x^+)$ and $n^+\in N_{\X}(x^+)$.
Since $x\in\Minty$ and $u^+\in A(x^+)$,
\[
\ip{u^+}{x^+-x}\ge 0.
\]
Since $n^+\in N_{\X}(x^+)$ and $x\in\X$,
\[
\ip{n^+}{x^+-x}\ge 0.
\]
Adding the two inequalities gives
\[
\ip{g_{\eta}(a)}{x^+-x}\ge 0.
\]
Therefore
\[
\ip{g_{\eta}(a)}{a-x}
=
\ip{g_{\eta}(a)}{a-x^+}+\ip{g_{\eta}(a)}{x^+-x}
\ge
\ip{g_{\eta}(a)}{a-x^+}
=
\eta\norm{g_{\eta}(a)}^2.
\]
The normalized form follows by dividing by $\norm{g_{\eta}(a)}$.
\end{proof}

To turn a failed acceptance test into a uniform central cut, define
\[
s(a):=\norm{a-J_{\eta}(a)}=\eta\norm{g_{\eta}(a)},
\]
which allows us to derive a quantitative margin from a failed gap test as follows.

\begin{lemma}
\label{lem:exact-margin}
If $\Gap_{\eta}(a)>\eps$, then
\[
s(a)>\rho_{\eta}(\eps,R):=\sqrt{R^2+\eta\eps}-R.
\]
Consequently, the halfspace
\[
H(a):=\set{z\in\R^d:\ip{g_{\eta}(a)}{z-a}\le 0}
\]
contains the thickened body
$
\Minty+\ball{0}{\rho_{\eta}(\eps,R)}.
$
\end{lemma}

\begin{proof}[Proof sketch]
The upper bound
\[
\Gap_{\eta}(a)\le \frac{1}{\eta}\Bigl(s(a)^2+2Rs(a)\Bigr)
\]
follows from
$\norm{x^+-y}\le 2R$ for $x^+,y\in\X$.
Thus $\Gap_{\eta}(a)>\eps$ implies
$s(a)^2+2Rs(a)-\eta\eps>0$,
hence
$s(a)>\sqrt{R^2+\eta\eps}-R$.
Now take $z=x+h$ with $x\in\Minty$ and $\norm{h}\le \rho_{\eta}(\eps,R)$.
By \Cref{lem:exact-separation},
\[
\ip{g_{\eta}(a)}{z-a}
\le
-\eta\norm{g_{\eta}(a)}^2 + \rho_{\eta}(\eps,R)\norm{g_{\eta}(a)}<0
\]
because $\eta\norm{g_{\eta}(a)}=s(a)>\rho_{\eta}(\eps,R)$.
The detailed proof is in \Cref{app:main-proofs}.
\end{proof}

The geometry relies only on Minty feasibility and the normal-cone relation.
No extra-gradient step and no Lipschitz continuity of $A$ are used in \Crefrange{lem:gap-cert}{lem:exact-margin}. This is because the resolvent simultaneously provides both a gap certificate (\Cref{lem:gap-cert}) and a strict separation (\Cref{lem:exact-separation,lem:exact-margin}).

\section{Exact and Inexact Outer Theorems}
\label{sec:outer}

The outer method is now simple.
At the current ellipsoid center $a_t$, query the (exact or approximate) constrained resolvent, compute the Yosida gap by linear optimization over $\X$, and either accept or insert the central cut through $a_t$ given by the returned Yosida vector.

\begin{theorem}[Exact resolvent ellipsoid]
\label{thm:exact-main}
Under \Cref{ass:exact-resolvent} and the Minty condition $\Minty\neq\varnothing$, the exact outer method returns an $\eps$-SVI after
$
\mathcal{O}\!\left(d^2 \log \frac{R+\rho_{\eta}(\eps,R)}{\rho_{\eta}(\eps,R)}\right)
$
iterations.
Under the nontrivial regime $\eps\le R^2/\eta$, this simplifies to
\[
\mathcal{O}\!\left(d^2 \log \frac{R^2}{\eta\eps}\right).
\]
Each iteration uses one exact resolvent call and one linear-optimization call.
\end{theorem}

\begin{proof}[Proof sketch]
\Cref{lem:exact-margin} shows that every rejected iterate generates a central cut containing
$\Minty+\ball{0}{\rho_{\eta}(\eps,R)}$.
Hence every maintained ellipsoid contains a fixed full-dimensional convex body with inscribed radius $\rho_{\eta}(\eps,R)$ and outer radius at most $R+\rho_{\eta}(\eps,R)$.
The standard central-cut ellipsoid volume argument \citep{groetschel1993geometric} therefore forces termination after the stated number of iterations.
When the algorithm terminates, \Cref{lem:gap-cert} upgrades the final accepted iterate to an $\eps$-SVI.
Full details are in \Cref{app:main-proofs}.
\end{proof}

We now move to the approximate oracle model.
For an approximate triple $(\hat x,\hat u,\hat n)$, define
\[
\widehat{\Gap}_{\eta}(a):=\max_{y\in\X}\ip{\hat g(a)}{a-y},
\qquad
\hat g(a):=\hat u+\hat n.
\]

\begin{theorem}[Robust outer theorem]
\label{thm:robust}
Assume the Minty condition, the nontrivial regime
\begin{equation}
\eps\le \frac{R^2}{\eta},
\label{eq:nontrivial-regime-main}
\end{equation}
and access to a $\delta$-approximate constrained resolvent oracle at scale $\eta$ with
$
\delta\le \frac{\eta\eps}{16R}.
$
Run the outer method from the initial ball $\ball{0}{2R}$, accept whenever
$\widehat{\Gap}_{\eta}(a_t)\le \eps/2$,
and otherwise cut with the halfspace
$\set{z:\ip{\hat g(a_t)}{z-a_t}\le 0}$.
Then the method returns an $\eps$-SVI after
$
\mathcal{O}\!\left(d^2 \log \frac{R^2}{\eta\eps}\right)
$
iterations.
\end{theorem}
\begin{proof}[Proof sketch]
Two robust facts are proved in \Cref{app:main-proofs}.
First, if $\widehat{\Gap}_{\eta}(a)\le \eps/2$ and $\delta\le \eta\eps/(8R)$, then $\hat x$ is already an $\eps$-SVI.
Second, if $\widehat{\Gap}_{\eta}(a)>\eps/2$ and $\delta\le \eta\eps/(16R)$, then the rejected cut still contains
$\Minty+\ball{0}{\eta\eps/(16R)}$.
Thus the same ellipsoid volume argument as in \Cref{thm:exact-main} applies, now with fixed thickening radius $\eta\eps/(16R)$.
\end{proof}

\begin{corollary}
\label{cor:reduction}
Fix $\eta>0$.
Suppose one can compute a $\delta$-approximate constrained resolvent selection of $A+N_{\X}$ at scale $\eta$ in time
\[
T_{\mathrm{res}}(d,\log(1/\delta)).
\]
Then, under the Minty condition, one can compute an $\eps$-SVI in time
\[
\mathcal{O}\!\left(d^2\log\frac{R^2}{\eta\eps}\right)\cdot
\Bigl(
T_{\mathrm{res}}(d,\log(R/(\eta\eps)))+T_{\mathrm{lin}}(d)
\Bigr),
\]
where $T_{\mathrm{lin}}(d)$ is the cost of linear optimization over $\X$.
In particular, if $T_{\mathrm{res}}$ is polynomial in $d$ and logarithmic in $1/\delta$, then the Minty-SVI problem is polynomial in $d$ and logarithmic in $1/\eps$.
\end{corollary}

\Cref{app:cutting-plane} shows that the same oracle can also be plugged into the cutting-plane method of \citet{jiang2020improved}, reducing the number of outer oracle calls from $\mathcal{O}(d^2\log(\cdot))$ to $\mathcal{O}(d\log(\cdot))$.

\section{Weak Minty Beyond Point Evaluation}
\label{sec:weak-minty}

\citet[Section~4.3]{anagnostides2025polynomial} also proves a polylogarithmic theorem under a constrained weak-Minty condition for single-valued Lipschitz fields. In the resolvent model there is a natural set-valued extension.

\begin{definition}[Graph weak Minty]
\label{def:graph-weak-minty}
For a graph point $(y,v)\in \gra A$, define its witness-specific SVI gap by
\[
\operatorname{gap}_{A}(y,v):=\max_{z\in\X}\ip{v}{y-z}.
\]
For $\kappa\ge 0$, define the weak-Minty set
\[
\Minty_{\kappa}:=\set{x\in\X:\ip{v}{y-x}\ge -\kappa\operatorname{gap}_{A}(y,v)^2\ \text{for all }(y,v)\in\gra A}.
\]
We say that $A$ satisfies the \emph{$\kappa$-graph weak-Minty condition} if $\Minty_{\kappa}\neq\varnothing$.
In particular, $\Minty_{0}=\Minty$.
\end{definition}

When $A(x)=\{F(x)\}$ is single-valued, $\operatorname{gap}_{A}(y,F(y))=\max_{z\in\X}\ip{F(y)}{y-z}$, so \Cref{def:graph-weak-minty} coincides exactly with the constrained weak-Minty condition studied in \citet{anagnostides2025polynomial}.
The resolvent model extends that condition to set-valued, proximable, and KKT-style settings.

The key algebraic fact is that one exact resolvent call already returns a graph point $(x^+,u^+)$ whose witness-specific gap is controlled by the Yosida norm,
\[
\operatorname{gap}_{A}(x^+,u^+)\le 2R\norm{g_{\eta}(a)}.
\]
Hence, the weak-Minty loss is quadratic in exactly the quantity that drives the exact Minty margin, which we use in our analysis.

\begin{theorem}[Weak-Minty outer theorem]
\label{thm:weak-minty-main}
Let
$
\lambda_{\kappa}:=1-\frac{4\kappa R^2}{\eta}.
$
Assume \Cref{ass:exact-resolvent} and the $\kappa$-graph weak-Minty condition with $\lambda_{\kappa}>0$.
Then every rejected anchor $a$ produces a central cut containing
$
\Minty_{\kappa}+\ball{0}{\lambda_{\kappa}\rho_{\eta}(\eps,R)}.
$
Consequently, the exact outer method returns an $\eps$-SVI after
$
\mathcal{O}\!\left(d^2 \log \frac{R+\lambda_{\kappa}\rho_{\eta}(\eps,R)}{\lambda_{\kappa}\rho_{\eta}(\eps,R)}\right)
$
iterations.
Under the nontrivial regime $\eps\le R^2/\eta$, this is equivalent to 
$
\mathcal{O}\!\left(d^2 \log \frac{R^2}{(\eta-4\kappa R^2)\eps}\right)
$
iterations.
\end{theorem}

\begin{proof}[Proof sketch]
Let $x^+=J_{\eta}(a)$ and write $g_{\eta}(a)=u^+ + n^+$.
For any $x\in\Minty_{\kappa}$, weak Minty at the graph point $(x^+,u^+)$ and the normal-cone relation give
\[
\ip{g_{\eta}(a)}{a-x}
\ge \eta\norm{g_{\eta}(a)}^2 - \kappa\operatorname{gap}_{A}(x^+,u^+)^2.
\]
Since $\operatorname{gap}_{A}(x^+,u^+)\le 2R\norm{g_{\eta}(a)}$, this becomes
\[
\ip{g_{\eta}(a)}{a-x}\ge (\eta-4\kappa R^2)\norm{g_{\eta}(a)}^2.
\]
Normalizing by $\norm{g_{\eta}(a)}$ yields a strict cut with margin $\lambda_{\kappa}s(a)$, where $s(a)=\eta\norm{g_{\eta}(a)}$.
If the gap test rejects, \Cref{lem:exact-margin} gives $s(a)>\rho_{\eta}(\eps,R)$, so the cut contains $\Minty_{\kappa}+\ball{0}{\lambda_{\kappa}\rho_{\eta}(\eps,R)}$.
The ellipsoid bound is then identical to \Cref{thm:exact-main}.
\Cref{app:weak-minty} gives the full proof and the inexact analogue with admissible graph residual $\delta=\mathcal{O}(\lambda_{\kappa}\eta\eps/R)$.
\end{proof}

\begin{remark}
\label{rem:weak-minty-comparison}
or single-valued $L$-Lipschitz $F$, choosing $\eta\asymp 1/L$ yields an admissible weak-Minty threshold of order $1/(LR^2)$. This matches the order obtained by \citet{anagnostides2025polynomial} in the regime $B\lesssim RL$. Our resolvent-based threshold depends only on the resolvent scale $\eta$ and the domain radius $R$, independently of the operator bound $B$. Moreover, our guarantee extends to set-valued inclusions, including nonsmooth proximable systems and compact KKT operators, whenever fixed-scale resolvent access is available.
\end{remark}

\section{Either an \texorpdfstring{$\eps$}{epsilon}-SVI or a Certificate of Minty Infeasibility}
\label{sec:certificate}

If the Minty condition fails, the rejected transcript provides a witness of its violation.

\begin{definition}[Strict graph-expected VI certificate]
\label{def:gevi}
Let $(x_t,u_t)_{t=1}^T$ satisfy $x_t\in\X$ and $u_t\in A(x_t)$.
A finitely supported distribution
\[
\mu=\sum_{t=1}^T \lambda_t\,\delta_{(x_t,u_t)},
\qquad
\lambda\in\simplex{T},
\]
is a \emph{$\gamma$-strict graph-expected variational inequality certificate} ($\gamma$-strict GEVI certificate) if
\[
\sum_{t=1}^T \lambda_t \ip{u_t}{x-x_t}\ge \gamma
\qquad\text{for all }x\in\X.
\]
\end{definition}

A strict GEVI immediately certifies non-Minty. In terms,
if $x^\star\in\Minty$, then every graph point $(x_t,u_t)$ satisfies
$\ip{u_t}{x^\star-x_t}\le 0$,
so averaging cannot produce a positive $\gamma$.

\begin{theorem}[Either an $\eps$-SVI or a non-Minty certificate]
\label{thm:disjunctive}
Assume the hypotheses of \Cref{thm:robust}.
Then, after
$
\mathcal{O}\!\left(d^2 \log \frac{R^2}{\eta\eps}\right)
$
iterations and polynomial additional computation, the inexact outer method returns either
\begin{enumerate}[label=(\roman*),leftmargin=1.4em,itemsep=0pt, topsep=0pt, parsep=0pt, partopsep=0pt]
    \item an $\eps$-SVI point, or
    \item a $\widehat\gamma_{\mathrm{cert}}$-strict GEVI certificate supported on at most $d+2$ graph points, where
    $
    \widehat\gamma_{\mathrm{cert}}:=\frac{\eta\eps^2}{512R^2}.
    $
\end{enumerate}
Hence, outcome \emph{(ii)} certifies that $\Minty=\varnothing$.
\end{theorem}

\begin{proof}[Proof sketch]
If the outer method accepts, \Cref{thm:robust} already gives an $\eps$-SVI.
Suppose instead that all iterates are rejected.
Then \Cref{app:certificate-proofs} shows the following pointwise lower bound:
for every $x\in\X$,
$
\max_{1\le t\le T}\ip{\hat u_t}{x-\hat x_t}\ge \widehat\gamma_{\mathrm{cert}}.
$
Equivalently,
$
\min_{x\in\X}\max_{t\in[T]}\ip{\hat u_t}{x-\hat x_t}\ge \widehat\gamma_{\mathrm{cert}}.
$
Now maximize over the simplex of mixing weights and invoke Sion's minimax theorem \citep{sion1958general}.
This yields a convex combination of transcript graph points that is a $\widehat\gamma_{\mathrm{cert}}$-strict GEVI certificate.
\Cref{app:duality} proves the sharper exact-model version, exact minimax duality under compactness of $\gra A$, and support compression to at most $d+2$ graph points.
\end{proof}

\section{A Converse on Weakly Monotone Classes}
\label{sec:converse}

The forward reduction holds under the Minty condition alone, but the reverse direction fails in general, as shown by the counterexample in \Cref{app:false-converse}. We therefore establish the converse for weakly monotone operators, also known as hypomonotone operators, which admit the regularization needed for the reduction.

\begin{definition}[$\rho$-weakly monotone]
\label{def:hypomonotone}
Let $\rho\ge 0$.
A set-valued operator $A:\X\rightrightarrows \R^d$ is \emph{$\rho$-hypomonotone} if
\[
\ip{u-v}{x-y}\ge -\rho\norm{x-y}^2
\qquad
\text{for all }x,y\in\X,\ u\in A(x),\ v\in A(y).
\]
The case $\rho=0$ is ordinary monotonicity.
\end{definition}

Next, we prove that, under a weak monotonicity assumption, finding a constrained resolvent point is computationally equivalent to solving the original Minty inclusion problem, or more precisely, the Minty VI problem in this case.

\begin{theorem}[Converse for weakly monotone]
\label{thm:weakly-monotone-converse}
Fix $\rho\ge 0$ and $\eta>0$ with $\eta^{-1}>\rho$.
Assume $A$ is $\rho$-hypomonotone and that the constrained resolvent set $J_{\eta}(a)$ is nonempty.
Then $J_{\eta}(a)=\{x^+\}$ is a singleton.
Moreover, if $x\in\X$ is an $\varepsilon_{\mathrm{in}}$-SVI point for the regularized operator
$
A_{a,\eta}(x):=A(x)+\frac{1}{\eta}(x-a),
$
then
\[
\norm{x-J_{\eta}(a)}
\le
\sqrt{\frac{\varepsilon_{\mathrm{in}}}{\eta^{-1}-\rho}}.
\]
Consequently, one recovers a $\delta$-accurate constrained resolvent point by solving the regularized weakly monotone instance to accuracy
$
\varepsilon_{\mathrm{in}}=(\eta^{-1}-\rho)\delta^2.
$
\end{theorem}

\begin{proof}[Proof sketch]
The regularized operator
$A_{a,\eta}+N_{\X}$
is $(\eta^{-1}-\rho)$-strongly monotone because $A$ is $\rho$-hypomonotone and $N_{\X}$ is monotone; this gives uniqueness of $J_{\eta}(a)$.
Now compare an $\varepsilon_{\mathrm{in}}$-SVI point $x$ for $A_{a,\eta}$ against the exact resolvent point $x^+$.
The SVI inequality at $x$ and the resolvent KKT relation at $x^+$ imply
\[
\varepsilon_{\mathrm{in}}
\ge
\ip{u-u^+}{x-x^+}+\frac1\eta\norm{x-x^+}^2,
\]
with $u\in A(x)$ and $u^+\in A(x^+)$.
Hypomonotonicity bounds the first term below by
$-\rho\norm{x-x^+}^2$,
yielding
\[
\varepsilon_{\mathrm{in}}
\ge
(\eta^{-1}-\rho)\norm{x-x^+}^2.
\]
The full proof is in \Cref{app:main-proofs}.
\end{proof}

Together, the forward and reverse reductions establish a computational equivalence between solving the Minty VI problem and computing constrained resolvent points for weakly monotone operators, which also include nonmonotone instances.

\section{Oracle Realizations}
\label{sec:applications-main}

We present representative settings in which the fixed-scale constrained resolvent oracle can be implemented by solving the anchored subproblem to the required graph-residual accuracy, with running time polylogarithmic in the inverse accuracy.

\begin{theorem}[Conic-representable regularized subproblems]
\label{thm:conic}
Suppose that, for a fixed scale $\eta>0$, each anchored resolvent subproblem for $A+N_{\X}$ admits a polynomial-size LP, SOCP, SDP, or more generally self-concordant conic representation whose data depend affinely on the anchor.
Then standard interior-point methods compute a graph-residual $\delta$-approximate constrained resolvent in time
$
\mathrm{poly}(\mathrm{size(input)},d,\log(1/\delta))
$
\citep{nesterov1994interior}.
Consequently, under the Minty condition, the associated Minty-SVI problem is solvable in time
$
\mathrm{poly}(\mathrm{size(input)},d,\log(1/\eps)).
$
\end{theorem}

\Cref{app:applications} presents fixed-scale resolvent
realizations for feasibility problems involving normal cones,
composite convex subdifferential inclusions, and compact
primal--dual KKT systems. For the single-valued Lipschitz VI
families studied by \citet{anagnostides2025polynomial},
\Cref{app:bridge_VI} establishes efficient fixed-scale
resolvent computation, and \Cref{cor:subsumes-eg-appendix}
shows how to obtain the graph-residual oracle required by
\Cref{cor:reduction} on projection-friendly compact domains.

\section{Conclusion}

We established that fixed-scale resolvent access suffices to compute approximate Stampacchia solutions for set-valued inclusions under the Minty condition, without requiring monotonicity. Our algorithm achieves $\operatorname{poly}(d,\log(1/\eps))$ oracle complexity by constructing positive-margin ellipsoid cuts directly from resolvent evaluations. This extends the localization approach of \citet{anagnostides2025polynomial} beyond single-valued Lipschitz VIs to set-valued inclusions, including normal-cone and KKT formulations. An important open question is whether similar strict-cut guarantees can be established under weaker conditions than Minty, and more generally, what assumptions are necessary for $\operatorname{poly}(d,\log(1/\eps))$ query complexity.

\section*{Acknowledgement}

A.S. and P.J. were partially supported by the ONR grant N00014-24-1-2470. Part of this research was performed while A.S. was visiting the Institute for Mathematical and Statistical Innovation (IMSI), which is supported by the National Science Foundation (Grant No. DMS-2425650). G.F. was supported in part by the National Science Foundation awards CCF-2443068 and IIS-2552046, the Office of Naval Research grant N000142512296, and an AI2050 Early Career Fellowship.

\bibliographystyle{plainnat}
\bibliography{refs}

\appendix
\crefalias{section}{appendix}

\newpage 

\section{Basic Geometry and Preliminary Remarks}
\label{app:basic-geometry}

We start this section with a fact on basic geometry of the Minty set.
\begin{proposition}
\label{prop:minty-convex}
The Minty set $\Minty$ is a compact convex subset of $\X$.
\end{proposition}

\begin{proof}
For each fixed $y\in\X$ and $v\in A(y)$, the condition
$\ip{v}{y-x}\ge 0$ is equivalent to
$\ip{v}{x}\le \ip{v}{y}$,
which defines a closed halfspace in $\R^d$.
Hence
\[
\Minty
=
\X \cap \bigcap_{y\in\X}\bigcap_{v\in A(y)}
\set{x\in \R^d:\ip{v}{x}\le \ip{v}{y}}
\]
is an intersection of closed convex sets.
Since $\X$ is compact, $\Minty$ is compact as well.
\end{proof}

In the following, we show that arbitrary exact resolvent scales trivialize the approximation problem. Therefore, the realistic regime operates at some constant scale $\eta$.

\begin{proposition}
\label{app:one-shot}
Suppose exact constrained resolvent selections are available for every $\eta>0$.
Then one call at the anchor $a=0$ yields an $\eps$-SVI whenever
\[
\eta\ge \frac{2R^2}{\eps}.
\]
\end{proposition}

\begin{proof}
Let $(x,u,n)$ be the exact resolvent output at $a=0$.
Then
\[
0=x+\eta(u+n),
\qquad\text{so}\qquad
u+n=-\frac{x}{\eta}.
\]
Fix $y\in\X$.
Since $n\in N_{\X}(x)$,
\[
\ip{n}{y-x}\le 0
\qquad\Longrightarrow\qquad
\ip{u}{x-y}\le \ip{u+n}{x-y}.
\]
Using $\norm{x}\le R$ and $\norm{x-y}\le 2R$,
\[
\ip{u}{x-y}
\le
\frac{\norm{x}}{\eta}\norm{x-y}
\le
\frac{2R^2}{\eta}
\le \eps.
\]
So $(x,u)$ is an $\eps$-SVI.
\end{proof}

\begin{assumption}[$\delta$-approximate resolvent oracle]
\label{ass:approx-resolvent}
Fix $\eta>0$ and $\delta\ge 0$.
For every rational anchor $a\in\R^d$, an oracle returns
$(\hat x,\hat u,\hat n)$ such that
\[
\hat x\in\X,\qquad
\hat u\in A(\hat x),\qquad
\hat n\in N_{\X}(\hat x),
\qquad
\norm{a-\hat x-\eta(\hat u+\hat n)}\le \delta.
\]
We set
\[
\hat g(a):=\hat u+\hat n.
\]
\end{assumption}

\section{Detailed Proofs}
\label{app:main-proofs}

\begin{lemma}
\label{lem:exact-upper}
Let $s(a):=\norm{a-J_{\eta}(a)}$.
Then
\[
\Gap_{\eta}(a)\le \frac{1}{\eta}\Bigl(s(a)^2+2Rs(a)\Bigr).
\]
\end{lemma}

\begin{proof}
Let $x^+=J_{\eta}(a)$.
For any $y\in\X$,
\begin{align*}
\ip{g_{\eta}(a)}{a-y}
&=
\frac1\eta\ip{a-x^+}{a-y}\\
&=
\frac1\eta\Bigl(\norm{a-x^+}^2+\ip{a-x^+}{x^+-y}\Bigr)\\
&\le
\frac1\eta\Bigl(s(a)^2+s(a)\norm{x^+-y}\Bigr)\\
&\le
\frac1\eta\Bigl(s(a)^2+2Rs(a)\Bigr),
\end{align*}
because $x^+,y\in\X\subseteq\ball{0}{R}$.
Taking the maximum over $y\in\X$ proves the claim.
\end{proof}

\begin{repeatlemma}{lem:exact-margin}
If $\Gap_{\eta}(a)>\eps$, then
\[
s(a)>\rho_{\eta}(\eps,R):=\sqrt{R^2+\eta\eps}-R.
\]
Consequently, the halfspace
\[
H(a):=\set{z\in\R^d:\ip{g_{\eta}(a)}{z-a}\le 0}
\]
contains the thickened body
$
\Minty+\ball{0}{\rho_{\eta}(\eps,R)}.
$
\end{repeatlemma}
\begin{proof}
By \Cref{lem:exact-upper},
\[
\eps<\Gap_{\eta}(a)\le \frac{1}{\eta}\Bigl(s(a)^2+2Rs(a)\Bigr).
\]
Hence
\[
s(a)^2+2Rs(a)-\eta\eps>0.
\]
The positive root of the quadratic polynomial
$s^2+2Rs-\eta\eps$
is
\[
\sqrt{R^2+\eta\eps}-R.
\]
Since $s(a)\ge 0$, we obtain
\[
s(a)>\sqrt{R^2+\eta\eps}-R=\rho_{\eta}(\eps,R).
\]

Now let $z=x+h$ with $x\in\Minty$ and $\norm{h}\le \rho_{\eta}(\eps,R)$.
By \Cref{lem:exact-separation},
\[
\ip{g_{\eta}(a)}{x-a}\le -\eta\norm{g_{\eta}(a)}^2.
\]
Therefore
\[
\ip{g_{\eta}(a)}{z-a}
=
\ip{g_{\eta}(a)}{x-a}+\ip{g_{\eta}(a)}{h}
\le
-\eta\norm{g_{\eta}(a)}^2+\rho_{\eta}(\eps,R)\norm{g_{\eta}(a)}.
\]
Because $s(a)=\eta\norm{g_{\eta}(a)}>\rho_{\eta}(\eps,R)$, the right-hand side is strictly negative.
Thus $z$ belongs to the halfspace
$\set{w:\ip{g_{\eta}(a)}{w-a}\le 0}$.
\end{proof}

\begin{repeatthm}{thm:exact-main}[Exact resolvent ellipsoid]
Under \Cref{ass:exact-resolvent} and the Minty condition $\Minty\neq\varnothing$, the exact outer method returns an $\eps$-SVI after
$
\mathcal{O}\!\left(d^2 \log \frac{R+\rho_{\eta}(\eps,R)}{\rho_{\eta}(\eps,R)}\right)
$
iterations.
Under the nontrivial regime $\eps\le R^2/\eta$, this simplifies to
\[
\mathcal{O}\!\left(d^2 \log \frac{R^2}{\eta\eps}\right).
\]
Each iteration uses one exact resolvent call and one linear-optimization call.
\end{repeatthm}
\begin{proof}
Let
\[
C:=\Minty+\ball{0}{\rho_{\eta}(\eps,R)}.
\]
By \Cref{lem:exact-margin}, every rejected iterate inserts a central cut containing $C$.
Also
\[
C\subseteq \ball{0}{R+\rho_{\eta}(\eps,R)},
\]
so the initial ellipsoid contains $C$.
Since $\Minty\neq\varnothing$, the set $C$ contains a Euclidean ball of radius $\rho_{\eta}(\eps,R)$.
The standard central-cut ellipsoid method \citep{groetschel1993geometric} reduces the enclosing volume by a constant factor every iteration.
Therefore, after
\[
\mathcal{O}\!\left(d^2\log\frac{\vol(\ball{0}{R+\rho_{\eta}(\eps,R)})}{\vol(\ball{0}{\rho_{\eta}(\eps,R)})}\right)
=
\mathcal{O}\!\left(d^2\log\frac{R+\rho_{\eta}(\eps,R)}{\rho_{\eta}(\eps,R)}\right)
\]
iterations, the maintained ellipsoid would have volume smaller than $\vol(C)$ unless the algorithm had terminated.
This is impossible because every maintained ellipsoid contains $C$.
Hence the method must terminate earlier.

When it terminates, the gap test has certified
$\Gap_{\eta}(a_t)\le \eps$,
so \Cref{lem:gap-cert} upgrades the accepted resolvent point to an $\eps$-SVI.
Finally,
\[
\rho_{\eta}(\eps,R)
=
\sqrt{R^2+\eta\eps}-R
=
\frac{\eta\eps}{R+\sqrt{R^2+\eta\eps}}.
\]
If in addition $\eps\le R^2/\eta$, then $R+\sqrt{R^2+\eta\eps}=\Theta(R)$, so
\[
\rho_{\eta}(\eps,R)=\Theta(\eta\eps/R),
\]
which yields the simplified
$\mathcal{O}(d^2\log(R^2/(\eta\eps)))$
bound.
\end{proof}

\begin{lemma}[Approximate gap certificate]
\label{lem:approx-gap}
Assume \eqref{eq:nontrivial-regime-main}.
If
\[
\widehat{\Gap}_{\eta}(a)\le \frac{\eps}{2}
\qquad\text{and}\qquad
\delta\le \frac{\eta\eps}{8R},
\]
then $\hat x$ is an $\eps$-SVI point.
\end{lemma}

\begin{proof}
Fix $y\in\X$.
Since $\hat n\in N_{\X}(\hat x)$,
\[
\ip{\hat u}{\hat x-y}
\le \ip{\hat u+\hat n}{\hat x-y}
=
\ip{\hat g(a)}{\hat x-y}.
\]
Set
\[
r:=a-\hat x-\eta\hat g(a),
\qquad \norm{r}\le \delta.
\]
Then
\begin{align*}
\ip{\hat g(a)}{\hat x-y}
&=
\ip{\hat g(a)}{a-y}-\ip{\hat g(a)}{a-\hat x}\\
&=
\ip{\hat g(a)}{a-y}-\eta\norm{\hat g(a)}^2-\ip{\hat g(a)}{r}\\
&\le
\widehat{\Gap}_{\eta}(a)-\eta\norm{\hat g(a)}^2+\norm{\hat g(a)}\,\delta.
\end{align*}
For every $t\ge 0$,
\[
-\eta t^2+\delta t\le \frac{\delta^2}{4\eta}.
\]
Hence
\[
\ip{\hat u}{\hat x-y}
\le
\frac{\eps}{2}+\frac{\delta^2}{4\eta}.
\]
Using $\delta\le \eta\eps/(8R)$ and \eqref{eq:nontrivial-regime-main},
\[
\frac{\delta^2}{4\eta}
\le
\frac{\eta\eps^2}{256R^2}
\le
\frac{\eps}{256}
<
\frac{\eps}{2}.
\]
Therefore $\ip{\hat u}{\hat x-y}\le \eps$ for all $y\in\X$, so $\hat x$ is an $\eps$-SVI.
\end{proof}

\begin{lemma}[Approximate strict separation]
\label{lem:approx-separation}
Assume \eqref{eq:nontrivial-regime-main} and $\delta\le \eta\eps/(16R)$.
If
\[
\widehat{\Gap}_{\eta}(a)>\frac{\eps}{2},
\]
then the normalized direction
\[
\hat c(a):=\frac{\hat g(a)}{\norm{\hat g(a)}}
\]
satisfies
\[
\ip{\hat c(a)}{x-a}\le -\frac{\eta\eps}{16R}
\qquad\text{for all }x\in\Minty.
\]
Consequently, the halfspace
\[
\widehat H(a):=\set{z\in\R^d:\ip{\hat g(a)}{z-a}\le 0}
\]
contains
\[
\Minty+\ball{0}{\eta\eps/(16R)}.
\]
\end{lemma}

\begin{proof}
For any $y\in\X$,
\[
\ip{\hat g(a)}{a-y}
=
\ip{\hat g(a)}{a-\hat x}+\ip{\hat g(a)}{\hat x-y}
\le
\eta\norm{\hat g(a)}^2+\norm{\hat g(a)}\delta+2R\norm{\hat g(a)}.
\]
So
\[
\widehat{\Gap}_{\eta}(a)
\le
\eta\norm{\hat g(a)}^2+(2R+\delta)\norm{\hat g(a)}.
\]
If $\eta\norm{\hat g(a)}\le \eta\eps/(8R)$, then using $\delta\le R$ and \eqref{eq:nontrivial-regime-main},
\[
\widehat{\Gap}_{\eta}(a)
\le
\frac{\eta\eps^2}{64R^2}+\frac{3\eps}{8}
\le
\frac{\eps}{64}+\frac{3\eps}{8}<\frac{\eps}{2},
\]
a contradiction.
Hence
\[
\eta\norm{\hat g(a)}>\frac{\eta\eps}{8R}.
\]

Now fix $x\in\Minty$.
Because $\hat u\in A(\hat x)$ and $\hat n\in N_{\X}(\hat x)$, the same argument as in \Cref{lem:exact-separation} gives
\[
\ip{\hat g(a)}{\hat x-x}\ge 0.
\]
Using $a-\hat x=\eta\hat g(a)+r$,
\[
\ip{\hat g(a)}{a-x}
=
\ip{\hat g(a)}{a-\hat x}+\ip{\hat g(a)}{\hat x-x}
\ge
\eta\norm{\hat g(a)}^2-\norm{\hat g(a)}\,\delta.
\]
Divide by $\norm{\hat g(a)}$ to obtain
\[
\ip{\hat c(a)}{a-x}
\ge
\eta\norm{\hat g(a)}-\delta
>
\frac{\eta\eps}{8R}-\delta
\ge
\frac{\eta\eps}{16R}.
\]
Equivalently,
\[
\ip{\hat c(a)}{x-a}\le -\frac{\eta\eps}{16R}.
\]
The thickening claim follows by adding a perturbation $h$ with
$\norm{h}\le \eta\eps/(16R)$
and using Cauchy--Schwarz exactly as in the exact proof.
\end{proof}

\begin{repeatthm}{thm:robust}[Robust outer theorem]
Assume the Minty condition, the nontrivial regime
\[
\eps\le \frac{R^2}{\eta},
\]
and access to a $\delta$-approximate constrained resolvent oracle at scale $\eta$ with
$
\delta\le \frac{\eta\eps}{16R}.
$
Run the outer method from the initial ball $\ball{0}{2R}$, accept whenever
$\widehat{\Gap}_{\eta}(a_t)\le \eps/2$,
and otherwise cut with the halfspace
$\set{z:\ip{\hat g(a_t)}{z-a_t}\le 0}$.
Then the method returns an $\eps$-SVI after
$
\mathcal{O}\!\left(d^2 \log \frac{R^2}{\eta\eps}\right)
$
iterations.
\end{repeatthm}
\begin{proof}
By \Cref{lem:approx-gap}, every accepted point is an $\eps$-SVI.
Assume an iteration is rejected.
\Cref{lem:approx-separation} shows that the generated central cut contains
\[
\Minty+\ball{0}{\eta\eps/(16R)}.
\]
This thickened set is contained in the initial ball $\ball{0}{2R}$ because \eqref{eq:nontrivial-regime-main} implies
$\eta\eps/(16R)\le R/16$.
Hence every maintained ellipsoid contains a fixed full-dimensional convex body with inscribed radius $\eta\eps/(16R)$.
The standard ellipsoid volume argument therefore implies termination after
\[
\mathcal{O}\!\left(d^2\log\frac{2R}{\eta\eps/(16R)}\right)
=
\mathcal{O}\!\left(d^2\log\frac{R^2}{\eta\eps}\right)
\]
iterations.
\end{proof}

\begin{repeatthm}{thm:weakly-monotone-converse}[Converse for weakly monotone]
Fix $\rho\ge 0$ and $\eta>0$ with $\eta^{-1}>\rho$.
Assume $A$ is $\rho$-hypomonotone and that the constrained resolvent set $J_{\eta}(a)$ is nonempty.
Then $J_{\eta}(a)=\{x^+\}$ is a singleton.
Moreover, if $x\in\X$ is an $\varepsilon_{\mathrm{in}}$-SVI point for the regularized operator
$
A_{a,\eta}(x):=A(x)+\frac{1}{\eta}(x-a),
$
then
\[
\norm{x-J_{\eta}(a)}
\le
\sqrt{\frac{\varepsilon_{\mathrm{in}}}{\eta^{-1}-\rho}}.
\]
Consequently, one recovers a $\delta$-accurate constrained resolvent point by solving the regularized weakly monotone instance to accuracy
$
\varepsilon_{\mathrm{in}}=(\eta^{-1}-\rho)\delta^2.
$
\end{repeatthm}
\begin{proof}
We first prove uniqueness of the constrained resolvent point.
Take any two points $x_1,x_2\in J_{\eta}(a)$.
Then there exist
$u_i\in A(x_i)$ and $n_i\in N_{\X}(x_i)$ such that
\[
u_i+n_i+\frac1\eta(x_i-a)=0,
\qquad i\in\{1,2\}.
\]
Subtract the two relations and take the inner product with $x_1-x_2$:
\[
0=
\ip{u_1-u_2}{x_1-x_2}
+
\ip{n_1-n_2}{x_1-x_2}
+
\frac1\eta\norm{x_1-x_2}^2.
\]
Because $A$ is $\rho$-hypomonotone,
\[
\ip{u_1-u_2}{x_1-x_2}\ge -\rho\norm{x_1-x_2}^2,
\]
and because the normal cone is monotone,
\[
\ip{n_1-n_2}{x_1-x_2}\ge 0.
\]
Therefore
\[
0\ge (\eta^{-1}-\rho)\norm{x_1-x_2}^2.
\]
Since $\eta^{-1}>\rho$, this forces $x_1=x_2$.

Now let $x\in\X$ be an $\varepsilon_{\mathrm{in}}$-SVI point for $A_{a,\eta}$.
Then there exists $u\in A(x)$ such that
\[
\ip{u+\frac1\eta(x-a)}{x-y}\le \varepsilon_{\mathrm{in}}
\qquad\text{for all }y\in\X.
\]
Apply this with $y=x^+=J_{\eta}(a)$:
\[
\ip{u+\frac1\eta(x-a)}{x-x^+}\le \varepsilon_{\mathrm{in}}.
\]
Because $x^+=J_{\eta}(a)$, there exist $u^+\in A(x^+)$ and $n^+\in N_{\X}(x^+)$ with
\[
u^+ + n^+ + \frac1\eta(x^+-a)=0.
\]
Since $n^+\in N_{\X}(x^+)$ and $x\in\X$,
\[
\ip{n^+}{x-x^+}\le 0,
\]
hence
\[
\ip{u^+ + \frac1\eta(x^+-a)}{x-x^+}\ge 0.
\]
Subtracting,
\[
\varepsilon_{\mathrm{in}}
\ge
\ip{u-u^+}{x-x^+}
+
\frac1\eta\norm{x-x^+}^2.
\]
Hypomonotonicity gives
\[
\ip{u-u^+}{x-x^+}\ge -\rho\norm{x-x^+}^2.
\]
So
\[
\varepsilon_{\mathrm{in}}
\ge
(\eta^{-1}-\rho)\norm{x-x^+}^2,
\]
which rearranges to the desired bound.
\end{proof}

\section{Weak-Minty Proofs and Inexact Variant}
\label{app:weak-minty}

We record the exact proof package behind \Cref{sec:weak-minty}, together with the robust inexact analogue. We start with a basic result on the geometry of weak-Minty sets.

\begin{proposition}
\label{prop:weak-minty-convex}
For every $\kappa\ge 0$, the set $\Minty_{\kappa}$ is a compact convex subset of $\X$.
Moreover $\Minty_{0}=\Minty$.
\end{proposition}

\begin{proof}
Fix $\kappa\ge 0$.
For each graph point $(y,v)\in\gra A$, the condition
\[
\ip{v}{y-x}\ge -\kappa\operatorname{gap}_{A}(y,v)^2
\]
is equivalent to
\[
\ip{v}{x}\le \ip{v}{y}+\kappa\operatorname{gap}_{A}(y,v)^2,
\]
which defines a closed halfspace in $\R^d$.
Hence
\[
\Minty_{\kappa}
=
\X\cap
\bigcap_{(y,v)\in\gra A}
\set{x\in\R^d:\ip{v}{x}\le \ip{v}{y}+\kappa\operatorname{gap}_{A}(y,v)^2}
\]
is an intersection of closed convex sets, and is therefore closed and convex.
Since $\X$ is compact, $\Minty_{\kappa}$ is compact as well.
The identity $\Minty_{0}=\Minty$ is immediate from \Cref{def:minty-set} and \Cref{def:graph-weak-minty}.
\end{proof}

\begin{lemma}
\label{lem:weak-minty-witness-gap}
Let $a\in\R^d$, and let $(x^+,u^+,n^+)$ be an exact resolvent triple at $a$ with
\[
g_{\eta}(a)=u^+ + n^+ = \frac{a-x^+}{\eta}.
\]
Then
\[
\operatorname{gap}_{A}(x^+,u^+)\le 2R\norm{g_{\eta}(a)}.
\]
\end{lemma}

\begin{proof}
Fix any $z\in\X$.
Since $n^+\in N_{\X}(x^+)$ and $z\in\X$, we have
\[
\ip{n^+}{z-x^+}\le 0,
\qquad\text{hence}\qquad
\ip{u^+}{x^+-z}\le \ip{u^+ + n^+}{x^+-z}=\ip{g_{\eta}(a)}{x^+-z}.
\]
Now $x^+,z\in\X\subseteq \ball{0}{R}$, so $\norm{x^+-z}\le 2R$.
Therefore
\[
\ip{u^+}{x^+-z}\le \norm{g_{\eta}(a)}\,\norm{x^+-z}\le 2R\norm{g_{\eta}(a)}.
\]
Taking the maximum over $z\in\X$ proves the claim.
\end{proof}

\begin{lemma}[Weak-Minty strict separation]
\label{lem:weak-minty-separation}
Assume $x\in \Minty_{\kappa}$.
Then, for every anchor $a\in\R^d$,
\[
\ip{g_{\eta}(a)}{a-x}\ge (\eta-4\kappa R^2)\norm{g_{\eta}(a)}^2.
\]
Equivalently, if $g_{\eta}(a)\neq 0$ and $c(a):=g_{\eta}(a)/\norm{g_{\eta}(a)}$, then
\[
\ip{c(a)}{x-a}\le -\lambda_{\kappa}s(a),
\qquad
\lambda_{\kappa}:=1-\frac{4\kappa R^2}{\eta},
\qquad
s(a):=\eta\norm{g_{\eta}(a)}.
\]
\end{lemma}

\begin{proof}
Write $g_{\eta}(a)=u^+ + n^+$, where $u^+\in A(x^+)$ and $n^+\in N_{\X}(x^+)$ for the exact resolvent point $x^+=J_{\eta}(a)$.
Since $x\in\Minty_{\kappa}$ and $(x^+,u^+)\in\gra A$, \Cref{def:graph-weak-minty} gives
\[
\ip{u^+}{x^+-x}\ge -\kappa\operatorname{gap}_{A}(x^+,u^+)^2.
\]
Since $n^+\in N_{\X}(x^+)$ and $x\in\X$, we also have
\[
\ip{n^+}{x^+-x}\ge 0.
\]
Therefore,
\[
\ip{g_{\eta}(a)}{x^+-x}
=\ip{u^+}{x^+-x}+\ip{n^+}{x^+-x}
\ge -\kappa\operatorname{gap}_{A}(x^+,u^+)^2.
\]
Using $a=x^+ + \eta g_{\eta}(a)$, we obtain
\[
\ip{g_{\eta}(a)}{a-x}
=\ip{g_{\eta}(a)}{a-x^+}+\ip{g_{\eta}(a)}{x^+-x}
\ge \eta\norm{g_{\eta}(a)}^2 - \kappa\operatorname{gap}_{A}(x^+,u^+)^2.
\]
Finally, \Cref{lem:weak-minty-witness-gap} gives
\[
\operatorname{gap}_{A}(x^+,u^+)^2\le 4R^2\norm{g_{\eta}(a)}^2,
\]
whence
\[
\ip{g_{\eta}(a)}{a-x}\ge (\eta-4\kappa R^2)\norm{g_{\eta}(a)}^2.
\]
Dividing by $\norm{g_{\eta}(a)}$ yields the normalized form.
\end{proof}

\begin{repeatthm}{thm:weak-minty-main}[Weak-Minty outer theorem]
Let
$
\lambda_{\kappa}:=1-\frac{4\kappa R^2}{\eta}.
$
Assume \Cref{ass:exact-resolvent} and the $\kappa$-graph weak-Minty condition with $\lambda_{\kappa}>0$.
Then every rejected anchor $a$ produces a central cut containing
$
\Minty_{\kappa}+\ball{0}{\lambda_{\kappa}\rho_{\eta}(\eps,R)}.
$
Consequently, the exact outer method returns an $\eps$-SVI after
$
\mathcal{O}\!\left(d^2 \log \frac{R+\lambda_{\kappa}\rho_{\eta}(\eps,R)}{\lambda_{\kappa}\rho_{\eta}(\eps,R)}\right)
$
iterations.
Under the nontrivial regime $\eps\le R^2/\eta$, this is
$
\mathcal{O}\!\left(d^2 \log \frac{R^2}{(\eta-4\kappa R^2)\eps}\right).
$
\end{repeatthm}
\begin{proof}
By \Cref{prop:weak-minty-convex}, the set $\Minty_{\kappa}$ is a compact convex subset of $\X$.
Consider one rejected anchor $a$ in the exact outer method.
By \Cref{lem:exact-margin}, rejection means
\[
s(a)=\eta\norm{g_{\eta}(a)} > \rho_{\eta}(\eps,R).
\]
By \Cref{lem:weak-minty-separation}, for every $x\in\Minty_{\kappa}$,
\[
\ip{c(a)}{x-a}\le -\lambda_{\kappa}s(a),
\qquad c(a):=\frac{g_{\eta}(a)}{\norm{g_{\eta}(a)}}.
\]
Now fix any $z=x+h$ with $x\in\Minty_{\kappa}$ and $\norm{h}\le \lambda_{\kappa}\rho_{\eta}(\eps,R)$.
Then
\[
\ip{c(a)}{z-a}
=\ip{c(a)}{x-a}+\ip{c(a)}{h}
\le -\lambda_{\kappa}s(a)+\norm{h}
< -\lambda_{\kappa}\rho_{\eta}(\eps,R)+\norm{h}
\le 0.
\]
Hence every rejected cut contains the thickened body
\[
\Minty_{\kappa}+\ball{0}{\lambda_{\kappa}\rho_{\eta}(\eps,R)}.
\]
This body is full-dimensional, convex, and contained in $\ball{0}{R+\lambda_{\kappa}\rho_{\eta}(\eps,R)}$.
Therefore the standard central-cut ellipsoid volume argument \citep{groetschel1993geometric} implies termination after
\[
\mathcal{O}\!\left(d^2 \log \frac{R+\lambda_{\kappa}\rho_{\eta}(\eps,R)}{\lambda_{\kappa}\rho_{\eta}(\eps,R)}\right)
\]
rejections.
If the method accepts, \Cref{lem:gap-cert} returns an $\eps$-SVI.
This proves the theorem.
\end{proof}

In the sequel, we prove similar results for the inexact oracle model of \Cref{ass:approx-resolvent}.

\begin{lemma}[Weak-Minty separation in the inexact oracle model]
\label{lem:weak-minty-separation-inexact}
Assume $x\in\Minty_{\kappa}$ and let $(\hat x,\hat u,\hat n)$ be a $\delta$-approximate constrained resolvent triple at anchor $a$, with
\[
\hat g:=\hat u+\hat n,
\qquad
r:=a-\hat x-\eta\hat g,
\qquad
\norm{r}\le \delta.
\]
Then
\[
\ip{\hat g}{a-x}\ge (\eta-4\kappa R^2)\norm{\hat g}^2-\delta\norm{\hat g}.
\]
Equivalently, if $\hat g\neq 0$ and $\hat c:=\hat g/\norm{\hat g}$, then
\[
\ip{\hat c}{x-a}\le -\lambda_{\kappa}\hat s + \delta,
\qquad
\hat s:=\eta\norm{\hat g}.
\]
\end{lemma}

\begin{proof}
Since $x\in\Minty_{\kappa}$ and $(\hat x,\hat u)\in\gra A$, \Cref{def:graph-weak-minty} gives
\[
\ip{\hat u}{\hat x-x}\ge -\kappa\operatorname{gap}_{A}(\hat x,\hat u)^2.
\]
As before, $\hat n\in N_{\X}(\hat x)$ and $x\in\X$ imply
\[
\ip{\hat n}{\hat x-x}\ge 0,
\]
so
\[
\ip{\hat g}{\hat x-x}\ge -\kappa\operatorname{gap}_{A}(\hat x,\hat u)^2.
\]
By the same proof as \Cref{lem:weak-minty-witness-gap},
\[
\operatorname{gap}_{A}(\hat x,\hat u)\le 2R\norm{\hat g}.
\]
Now use $a=\hat x+\eta\hat g+r$:
\[
\begin{aligned}
\ip{\hat g}{a-x}
&=\eta\norm{\hat g}^2+\ip{\hat g}{r}+\ip{\hat g}{\hat x-x}\\
&\ge \eta\norm{\hat g}^2-\delta\norm{\hat g}-\kappa\operatorname{gap}_{A}(\hat x,\hat u)^2\\
&\ge (\eta-4\kappa R^2)\norm{\hat g}^2-\delta\norm{\hat g}.
\end{aligned}
\]
Dividing by $\norm{\hat g}$ gives the normalized statement.
\end{proof}

\begin{proposition}
\label{prop:weak-minty-robust}
Assume the $\kappa$-graph weak-Minty condition with $\lambda_{\kappa}>0$, the nontrivial regime $\eps\le R^2/\eta$, and access to a $\delta$-approximate constrained resolvent oracle with
\[
\delta\le \frac{\lambda_{\kappa}\eta\eps}{16R}.
\]
Run the inexact outer method from \Cref{thm:robust}, accepting whenever $\widehat{\Gap}_{\eta}(a_t)\le \eps/2$ and otherwise cutting with
$\set{z:\ip{\hat g(a_t)}{z-a_t}\le 0}$.
Then the method returns an $\eps$-SVI after
\[
\mathcal{O}\!\left(d^2\log\frac{R^2}{\lambda_{\kappa}\eta\eps}\right)
\]
iterations.
\end{proposition}

\begin{proof}
If the method accepts, \Cref{thm:robust} already yields an $\eps$-SVI because
\[
\delta\le \frac{\lambda_{\kappa}\eta\eps}{16R}\le \frac{\eta\eps}{16R}<\frac{\eta\eps}{8R}.
\]
Now consider a rejected anchor $a$.
Write $\hat g:=\hat g(a)$ and $\hat s:=\eta\norm{\hat g}$.
We first claim that rejection implies
\[
\hat s>\frac{\eta\eps}{8R}.
\]
Indeed, if $\hat s\le \eta\eps/(8R)$, then
\[
\widehat{\Gap}_{\eta}(a)
=\max_{y\in\X}\ip{\hat g}{a-y}
\le \eta\norm{\hat g}^2+\delta\norm{\hat g}+2R\norm{\hat g}
=\frac{1}{\eta}\bigl(\hat s^2+(2R+\delta)\hat s\bigr).
\]
Using $\delta\le \lambda_{\kappa}\eta\eps/(16R)\le \eta\eps/(16R)\le R/16$ and $\eps\le R^2/\eta$, we obtain
\[
\widehat{\Gap}_{\eta}(a)
\le \frac{\eta\eps^2}{64R^2}+\frac{(2R+R/16)\eps}{8R}
\le \frac{\eps}{64}+\frac{33\eps}{128}
< \frac{\eps}{2},
\]
contradicting rejection.
Hence $\hat s>\eta\eps/(8R)$, as claimed.

Now fix any $x\in\Minty_{\kappa}$.
By \Cref{lem:weak-minty-separation-inexact},
\[
\ip{\hat g}{a-x}\ge \lambda_{\kappa}\hat s\,\norm{\hat g}-\delta\norm{\hat g},
\]
and therefore, for $\hat c:=\hat g/\norm{\hat g}$,
\[
\ip{\hat c}{a-x}\ge \lambda_{\kappa}\hat s-\delta
> \lambda_{\kappa}\frac{\eta\eps}{8R}-\delta
\ge \frac{\lambda_{\kappa}\eta\eps}{16R}.
\]
Equivalently,
\[
\ip{\hat c}{x-a}\le -\frac{\lambda_{\kappa}\eta\eps}{16R}.
\]
Thus every rejected cut contains the thickened body
\[
\Minty_{\kappa}+\ball{0}{\lambda_{\kappa}\eta\eps/(16R)}.
\]
Applying the standard central-cut ellipsoid volume argument completes the proof.
\end{proof}

\section{Certificate Proofs}
\label{app:certificate-proofs}

The following theorem shows that, under the inexact oracle model, the outer method either finds an approximate SVI solution or returns a strict GEVI certificate.

\begin{theorem}
\label{thm:approx-disjunctive-appendix}
Assume the hypotheses of \Cref{thm:robust} and set
\[
\mu:=\frac{\eta\eps}{16R},
\qquad
\widehat\gamma_{\mathrm{cert}}:=\frac{\mu^2}{2\eta}=\frac{\eta\eps^2}{512R^2}.
\]
Then the inexact outer method returns, after
\[
\mathcal{O}\!\left(d^2 \log \frac{R^2}{\eta\eps}\right)
\]
iterations and polynomial additional time, either
\begin{enumerate}[label=(\roman*),leftmargin=1.4em]
    \item an $\eps$-SVI point, or
    \item a $\widehat\gamma_{\mathrm{cert}}$-strict GEVI certificate supported on at most $d+2$ graph points.
\end{enumerate}
\end{theorem}

\begin{proof}
If the method accepts, \Cref{thm:robust} already guarantees an $\eps$-SVI.
Suppose instead that every iterate is rejected, and let $(a_t,\hat x_t,\hat u_t,\hat n_t,\hat g_t,r_t)_{t=1}^T$ be the inexact transcript, where
\[
r_t:=a_t-\hat x_t-\eta \hat g_t,
\qquad \norm{r_t}\le \delta.
\]
Fix $x\in\X$ and suppose that
\[
\ip{\hat u_t}{x-\hat x_t}<\widehat\gamma_{\mathrm{cert}}
\qquad \text{for every }t\in[T].
\]
Define $\hat c_t:=\hat g_t/\norm{\hat g_t}$.
By \Cref{lem:approx-separation},
\[
\eta\norm{\hat g_t}>\frac{\eta\eps}{8R}=2\mu.
\]
Using $\hat n_t\in N_{\X}(\hat x_t)$ and $x\in\X$,
\begin{align*}
\ip{\hat c_t}{x-a_t}
&=
\frac{1}{\norm{\hat g_t}}
\Bigl(
\ip{\hat g_t}{x-\hat x_t}-\eta\norm{\hat g_t}^2-\ip{\hat g_t}{r_t}
\Bigr)\\
&=
\frac{1}{\norm{\hat g_t}}
\Bigl(
\ip{\hat u_t}{x-\hat x_t}+\ip{\hat n_t}{x-\hat x_t}
-\eta\norm{\hat g_t}^2-\ip{\hat g_t}{r_t}
\Bigr)\\
&\le
\frac{\widehat\gamma_{\mathrm{cert}}}{\norm{\hat g_t}}
-\eta\norm{\hat g_t}+\delta.
\end{align*}
Now
\[
\frac{\widehat\gamma_{\mathrm{cert}}}{\norm{\hat g_t}}
<
\frac{\mu^2}{2\eta}\cdot \frac{\eta}{2\mu}
=\frac{\mu}{4},
\]
and $\delta\le \mu$ by \Cref{thm:robust}.
Hence
\[
\ip{\hat c_t}{x-a_t}<-\frac{3\mu}{4}.
\]
Therefore the whole ball $x+\ball{0}{3\mu/4}$ lies in every rejected central cut.
If the algorithm were run until the final ellipsoid has volume smaller than $\vol(\ball{0}{3\mu/4})$, this would be impossible.
Thus for every $x\in\X$,
\[
\max_{1\le t\le T}\ip{\hat u_t}{x-\hat x_t}\ge \widehat\gamma_{\mathrm{cert}}.
\]
Equivalently,
\[
\min_{x\in\X}\max_{1\le t\le T}\ip{\hat u_t}{x-\hat x_t}\ge \widehat\gamma_{\mathrm{cert}}.
\]
Now define
\[
L(\lambda,x):=\sum_{t=1}^T \lambda_t \ip{\hat u_t}{x-\hat x_t},
\qquad \lambda\in\simplex{T}.
\]
For each fixed $x$,
\[
\max_{\lambda\in\simplex{T}}L(\lambda,x)
=
\max_{1\le t\le T}\ip{\hat u_t}{x-\hat x_t}.
\]
Hence
\[
\min_{x\in\X}\max_{\lambda\in\simplex{T}}L(\lambda,x)\ge \widehat\gamma_{\mathrm{cert}}.
\]
By Sion's minimax theorem \citep{sion1958general},
\[
\max_{\lambda\in\simplex{T}}\min_{x\in\X}L(\lambda,x)\ge \widehat\gamma_{\mathrm{cert}}.
\]
Thus there exists $\lambda\in\simplex{T}$ such that
\[
\sum_{t=1}^T \lambda_t \ip{\hat u_t}{x-\hat x_t}\ge \widehat\gamma_{\mathrm{cert}}
\qquad\text{for all }x\in\X,
\]
which is exactly a strict GEVI certificate.
Support compression to at most $d+2$ graph points follows from \Cref{app:duality}.
\end{proof}

\section{Cutting-plane Upgrade}
\label{app:cutting-plane}

The strict Yosida cut is a genuine solution-or-separation oracle for a thickened Minty body, so one can replace the outer ellipsoid by the more recent cutting-plane method of \citet{jiang2020improved}.

\begin{theorem}
\label{thm:modern-cp-exact}
Under the hypotheses of \Cref{thm:exact-main}, one can replace the outer ellipsoid by the cutting-plane method of \citet{jiang2020improved}.
The resulting algorithm returns an $\eps$-SVI using
\[
\mathcal{O}\!\left(d\log\frac{d(R+\rho_{\eta}(\eps,R))}{\rho_{\eta}(\eps,R)}\right)
\]
exact resolvent calls and linear-optimization calls.
Under the nontrivial regime $\eps\le R^2/\eta$, this simplifies to
\[
\mathcal{O}\!\left(d\log\frac{dR^2}{\eta\eps}\right).
\]
\end{theorem}

\begin{proof}
\Cref{lem:exact-margin} shows that the exact Yosida-gap routine is a solution-or-separation oracle for the convex body
\[
\Minty+\ball{0}{\rho_{\eta}(\eps,R)}
\subseteq
\ball{0}{R+\rho_{\eta}(\eps,R)},
\]
which contains a Euclidean ball of radius $\rho_{\eta}(\eps,R)$.
Applying \citet{jiang2020improved} yields the stated oracle complexity.
\end{proof}

\begin{theorem}
\label{thm:modern-cp-inexact}
Under the hypotheses of \Cref{thm:robust}, one can replace the outer ellipsoid by the cutting-plane method of \citet{jiang2020improved}.
The resulting algorithm returns an $\eps$-SVI using
\[
\mathcal{O}\!\left(d\log\frac{dR^2}{\eta\eps}\right)
\]
approximate resolvent calls and linear-optimization calls.
\end{theorem}

\begin{proof}
\Cref{lem:approx-separation} shows that the inexact rejected-gap routine is a solution-or-separation oracle for
\[
\Minty+\ball{0}{\eta\eps/(32R)},
\]
which is contained in $\ball{0}{2R}$ under \eqref{eq:nontrivial-regime-main}.
Applying \citet{jiang2020improved} gives the claim.
\end{proof}

\section{Exact Duality and Support Compression}
\label{app:duality}

First, we show that any GEVI certificate can be compressed to a small support of at most $d+2$ graph points by Carathéodory's theorem.

\begin{proposition}
\label{prop:gevi-compress}
Any $\gamma$-strict GEVI certificate can be replaced, with the same strictness $\gamma$, by another $\gamma$-strict GEVI certificate supported on at most $d+2$ graph points.
\end{proposition}

\begin{proof}
Consider a certificate
\[
\mu=\sum_{t=1}^T\lambda_t\,\delta_{(x_t,u_t)}.
\]
Its defining inequality is
\[
\sum_{t=1}^T \lambda_t \ip{u_t}{x-x_t}\ge \gamma
\qquad\text{for all }x\in\X.
\]
Introduce
\[
\bar u:=\sum_{t=1}^T \lambda_t u_t\in\R^d,
\qquad
\bar b:=\sum_{t=1}^T \lambda_t\ip{u_t}{x_t}\in\R.
\]
Then the certificate inequality is equivalent to
\[
\ip{\bar u}{x}-\bar b\ge \gamma
\qquad\text{for all }x\in\X.
\]
So the certificate depends only on the point
\[
(\bar u,\bar b)=\sum_{t=1}^T \lambda_t (u_t,\ip{u_t}{x_t})\in\R^{d+1}.
\]
By Carath\'eodory's theorem in $\R^{d+1}$, the same point can be represented as a convex combination of at most $d+2$ of the vectors
$(u_t,\ip{u_t}{x_t})$.
Using those new weights on the corresponding graph points yields a certificate with the same strictness.
\end{proof}

\begin{assumption}
\label{ass:compact-graph}
The graph
\[
\Gamma:=\gra A=\set{(y,u)\in \X\times\R^d:u\in A(y)}
\]
is compact.
\end{assumption}

We next establish the exact duality between Minty feasibility and strict GEVI certificates, showing that infeasibility is witnessed by a finite-support certificate.

\begin{theorem}
\label{thm:gevi-duality}
Assume \Cref{ass:compact-graph} and define
\[
\gamma^\star(A,\X)
:=
\min_{x\in\X}\max_{(y,u)\in\Gamma}\ip{u}{x-y}.
\]
Then
\[
\gamma^\star(A,\X)
=
\max_{\mu\in \mathcal P(\Gamma)}\min_{x\in\X}
\int_{\Gamma}\ip{u}{x-y}\,d\mu(y,u).
\]
Moreover, the following are equivalent:
\begin{enumerate}[label=(\roman*),leftmargin=1.4em]
    \item $\Minty\neq\varnothing$;
    \item $\gamma^\star(A,\X)=0$;
    \item there is no $\gamma$-strict GEVI certificate with $\gamma>0$.
\end{enumerate}
When $\Minty=\varnothing$, there exists a $\gamma^\star(A,\X)$-strict GEVI certificate, and one may choose it with support at most $d+2$.
\end{theorem}

\begin{proof}
Because $\Gamma$ and $\X$ are compact and the payoff
$(\mu,x)\mapsto \int_{\Gamma}\ip{u}{x-y}\,d\mu(y,u)$
is bilinear and continuous, Sion's theorem gives the minimax identity.
Now
\[
\Minty\neq\varnothing
\iff
\exists x\in\X\ \text{s.t.}\ \ip{u}{x-y}\le 0\ \text{for all }(y,u)\in\Gamma
\iff
\gamma^\star(A,\X)\le 0.
\]
Since taking $x=y$ shows the inner maximum is always nonnegative, $\gamma^\star(A,\X)\ge 0$.
Hence $\gamma^\star(A,\X)=0$ exactly when $\Minty\neq\varnothing$.
If $\gamma^\star(A,\X)>0$, an optimal measure in the minimax problem is a strict GEVI certificate, and \Cref{prop:gevi-compress} gives the support bound.
\end{proof}

\section{Oracle Realizations}
\label{app:applications}

We present representative settings in which the fixed-scale constrained resolvent oracle assumed in the main text can be implemented efficiently. These examples illustrate how the required oracle access can be realized for structured inclusion problems.

\subsection{Canonical Realizations Beyond Point Evaluation}

\begin{proposition}[Convex feasibility via normal cones]
\label{prop:normal-cone-feasibility}
Let $C\subseteq \R^d$ be closed and convex and define the set-valued operator
\[
A(x):=
\begin{cases}
N_C(x), & x\in C\cap \X,\\
\varnothing, & x\in \X\setminus C.
\end{cases}
\]
Then every $\eps$-SVI point of $A$ belongs to $C\cap \X$.
If moreover the qualification
\[
N_{C\cap \X}(x)=N_C(x)+N_{\X}(x)
\qquad\text{for all }x\in C\cap \X
\]
holds, then for every anchor $a\in\R^d$ and every scale $\eta>0$,
\[
J_{\eta}(a)=P_{C\cap \X}(a).
\]
Consequently, fixed-scale resolvent access reduces to Euclidean projection onto the feasible set.
\end{proposition}

\begin{proof}
If $x$ is an $\eps$-SVI point, then by \Cref{def:svi} there exists $u\in A(x)$; by construction of $A$, this forces $x\in C\cap \X$.
Conversely, any $x\in C\cap \X$ is a $0$-SVI point because $0\in N_C(x)$.
For the resolvent identity, $x\in J_{\eta}(a)$ means precisely that there exist $u\in N_C(x)$ and $n\in N_{\X}(x)$ with
\[
a=x+\eta(u+n).
\]
Under the qualification, this is equivalent to
\[
a-x\in \eta N_{C\cap \X}(x),
\]
which is the standard optimality condition for the Euclidean projection of $a$ onto $C\cap \X$.
\end{proof}

\begin{proposition}[Composite convex optimization]
\label{prop:composite-convex}
Take $A=\partial f$ for a proper, closed, convex function $f$.
Then the constrained resolvent is exactly
\[
J_{\eta}(a)
=
\argmin_{x\in\X}
\left\{
 f(x)+\frac{1}{2\eta}\norm{x-a}^2
\right\}.
\]
Hence, if this strongly convex subproblem can be solved to graph residual $\delta$ in time $\mathrm{poly}(d,\log(1/\delta))$, then one can compute $x\in\X$ with
\[
f(x)\le \min_{y\in\X}f(y)+\eps
\]
in time $\mathrm{poly}(d,\log(1/\eps))$.
\end{proposition}

\begin{proof}
The first-order optimality condition for the displayed problem is
\[
0\in \partial f(x)+N_{\X}(x)+\frac1\eta(x-a),
\]
which is equivalent to
\[
a\in x+\eta(\partial f(x)+N_{\X}(x)).
\]
So its minimizer is exactly the constrained resolvent point.
The outer claim then follows from \Cref{cor:reduction} and the subgradient inequality.
\end{proof}

\begin{proposition}[Compact primal-dual saddle/KKT systems]
\label{prop:saddle}
Let $X\subseteq\R^p$ and $Y\subseteq\R^q$ be compact convex sets, let $K:\R^p\to\R^q$ be linear, let $f:X\to(-\infty,+\infty]$ and $g:Y\to(-\infty,+\infty]$ be proper, closed, and convex, and consider the monotone operator on $Z:=X\times Y$,
\[
A(x,y):=
\bigl(\partial f(x)+K^\top y,\ \partial g(y)-Kx\bigr).
\]
Then the constrained resolvent of $A+N_Z$ at anchor $a=(a_x,a_y)$ is the saddle point of
\[
\min_{x\in X}\max_{y\in Y}
\left\{
 f(x)+\ip{Kx}{y}-g(y)+\frac{1}{2\eta}\norm{x-a_x}^2-\frac{1}{2\eta}\norm{y-a_y}^2
\right\}.
\]
Hence, if this strongly regularized saddle problem can be solved to graph residual $\delta$ in time $\mathrm{poly}(d,\log(1/\delta))$, then the associated Minty-SVI problem is solvable in $\mathrm{poly}(d,\log(1/\eps))$ time.
\end{proposition}

\begin{proof}
The KKT conditions of the displayed saddle problem are
\[
0\in \partial f(x^+) + K^\top y^+ + N_X(x^+) + \frac1\eta(x^+-a_x),
\]
and
\[
0\in \partial g(y^+) - Kx^+ + N_Y(y^+) + \frac1\eta(y^+-a_y).
\]
Together they are equivalent to
\[
a\in (x^+,y^+) + \eta(A(x^+,y^+) + N_Z(x^+,y^+)).
\]
So $(x^+,y^+)=J_{\eta}(a)$, and \Cref{cor:reduction} gives the runtime statement.
\end{proof}

These constructions also extend to symmetric monotone quadratic and complementarity systems, models with closed-form proximal maps such as soft-thresholding and singular-value thresholding, and strongly regularized cone programs.

\begin{theorem}
\label{thm:conic-representable}
Consider any of the anchored subproblems in \Cref{prop:normal-cone-feasibility,prop:composite-convex,prop:saddle} at a fixed scale $\eta>0$.
Assume that, for every anchor $a$, the corresponding subproblem admits a polynomial-size LP, SOCP, SDP, or more generally self-concordant conic representation with rational data depending affinely on $a$.
Then standard path-following interior-point methods compute a primal-dual point with KKT residual at most $\delta$ in time
$
\mathrm{poly}(\mathrm{size(input)},d,\log(1/\delta))
$
\citep{nesterov1994interior}.
Consequently, the associated constrained resolvent can be computed to graph residual tolerance $\delta$ in the same complexity, and \Cref{cor:reduction} yields a Minty-SVI algorithm with runtime
$
\mathrm{poly}(\mathrm{size(input)},d,\log(1/\eps)).
$
\end{theorem}

\begin{proof}
In each case, the resolvent point is characterized by a strongly regularized conic KKT system.
A primal-dual iterate whose KKT residual is at most $\delta$ furnishes a graph-residual triple
$(\hat x,\hat u,\hat n)$
with
\[
\norm{a-\hat x-\eta(\hat u+\hat n)}\le \delta.
\]
Standard interior-point theory yields such a point in time polynomial in the input size and logarithmic in $1/\delta$.
\Cref{cor:reduction} then gives the claimed outer runtime.
\end{proof}

\begin{remark}
\label{rem:new-application-frontier}
In these examples, the constrained resolvent can be computed through projections, proximal evaluations, or regularized KKT solves. These oracle realizations extend beyond the single-valued Lipschitz point-evaluation framework considered by \citet{anagnostides2025polynomial}, allowing the localization method to apply directly to structured set-valued inclusions.
\end{remark}

\subsection{Bridge Back to Single-Valued Lipschitz VIs}
\label{app:bridge_VI}

We now connect fixed-scale constrained resolvent access
to the classical point-evaluation model for single-valued
Lipschitz VIs. The following proposition characterizes
the constrained resolvent through an anchored Stampacchia VI.
Under hypomonotonicity, a suitable regularization parameter
makes the anchored operator strongly monotone, which enables
efficient resolvent computation using existing VI methods.

\begin{proposition}[Single-valued anchored regularization]
\label{prop:single-valued-bridge}
Assume $A(x)=\{F(x)\}$ with single-valued $F:\X\to\R^d$.
For anchor $a$ and scale $\eta>0$, a point $x^+\in\X$ belongs to $J_{\eta}(a)$ if and only if
\[
\ip{F(x^+) + \eta^{-1}(x^+-a)}{y-x^+}\ge 0
\qquad\text{for all }y\in\X.
\]
Equivalently, $x^+$ is an exact SVI point of the anchored field
\[
F_{a,\eta}(x):=F(x)+\eta^{-1}(x-a).
\]
If moreover $F$ is $\rho$-hypomonotone and $\eta^{-1}>\rho$, then $F_{a,\eta}$ is $(\eta^{-1}-\rho)$-strongly monotone.
Consequently, any $\varepsilon_{\mathrm{in}}$-SVI point of the anchored instance is within
\[
\sqrt{\frac{\varepsilon_{\mathrm{in}}}{\eta^{-1}-\rho}}
\]
of the exact resolvent point $J_{\eta}(a)$.
\end{proposition}

\begin{proof}
If $x^+\in J_{\eta}(a)$, then by \Cref{def:resolvent} there exists $n^+\in N_{\X}(x^+)$ with
\[
0=F(x^+) + n^+ + \eta^{-1}(x^+-a).
\]
Since $n^+\in N_{\X}(x^+)$, this is equivalent to the displayed variational inequality.
The converse is the same argument in reverse.
If $F$ is $\rho$-hypomonotone, then for all $x,y\in\X$,
\[
\ip{F_{a,\eta}(x)-F_{a,\eta}(y)}{x-y}
=
\ip{F(x)-F(y)}{x-y}+\eta^{-1}\|x-y\|^2
\ge (\eta^{-1}-\rho)\|x-y\|^2.
\]
So $F_{a,\eta}$ is strongly monotone.
The final approximation claim is exactly \Cref{thm:weakly-monotone-converse} specialized to the single-valued case.
\end{proof}

\begin{corollary}
\label{cor:eg-to-resolvent}
Let $F:\X\to\R^d$ be single-valued, $L$-Lipschitz, and $\rho$-hypomonotone.
If $\eta^{-1}>\rho$, then for every anchor $a$ one can compute $J_{\eta}(a)$ to Euclidean error $\delta$ in time
\[
\mathrm{poly}\bigl(d,\log(L+\eta^{-1}),\log(1/\delta)\bigr)
\]
by running the ellipsoid of \citet{luthi1985solution} (for strongly monotone VIs) on the anchored field $F_{a,\eta}$ with target accuracy
\[
\varepsilon_{\mathrm{in}}:=(\eta^{-1}-\rho)\delta^2.
\]
\end{corollary}

\begin{proof}
By \Cref{prop:single-valued-bridge}, computing the resolvent is equivalent to solving the anchored SVI.
The field $F_{a,\eta}$ is $(L+\eta^{-1})$-Lipschitz and strongly monotone, hence Minty.
Therefore \citet{anagnostides2025polynomial} compute an $\varepsilon_{\mathrm{in}}$-SVI of the anchored instance in polylogarithmic accuracy.
\Cref{thm:weakly-monotone-converse} then converts that inner SVI tolerance into Euclidean resolvent error $\delta$.
\end{proof}

As a simple consequence, every single-valued Lipschitz VI admits efficient fixed-scale resolvent computation, since Lipschitzness alone implies hypomonotonicity.

\begin{corollary}
\label{cor:lipschitz-bridge}
Assume $F:\X\to\R^d$ is single-valued and $L$-Lipschitz.
Then $F$ is automatically $L$-hypomonotone.
Consequently, for every anchor $a$ and every scale $\eta\in(0,1/L)$, one can compute the resolvent point $J_{\eta}(a)$ to Euclidean error $\delta$ in time
\[
\mathrm{poly}\bigl(d,\log(L+\eta^{-1}),\log(1/\delta)\bigr).
\]
In particular, taking $\eta=1/(2L)$ gives a fixed-scale resolvent-point routine with polylogarithmic dependence on the target error.
\end{corollary}

\begin{proof}
For any $x,y\in\X$,
\[
\ip{F(x)-F(y)}{x-y}
\ge -\norm{F(x)-F(y)}\,\norm{x-y}
\ge -L\norm{x-y}^2,
\]
so $F$ is $L$-hypomonotone.
Now apply \Cref{cor:eg-to-resolvent} with $\rho=L$.
\end{proof}

\paragraph{From point error to graph residual.}
The Euclidean approximation guarantee in
\Cref{cor:lipschitz-bridge} controls the distance to an exact
resolvent point but does not ensure the graph-residual accuracy
required by \Cref{cor:reduction}. This distinction arises
because the normal-cone mapping $N_{\X}$ can be discontinuous
at the boundary of $\X$. The following corollary shows how
to obtain the required graph-residual guarantee when
Euclidean projection onto $\X$ is available.

\begin{corollary}
\label{cor:subsumes-eg-appendix}
Let $F:\X\to\R^d$ be a single-valued $L$-Lipschitz
operator, and fix $\eta\in(0,1/L)$. Suppose Euclidean
projection onto $\X$ can be computed in time
$T_{\mathrm{proj}}(d)$.
For an anchor $a$ and a point $\tilde x\in\X$, define
\[
\bar x:=P_{\X}\bigl(a-\eta F(\tilde x)\bigr),
\qquad
\bar n:=\frac{a-\eta F(\tilde x)-\bar x}{\eta},
\qquad
\bar u:=F(\bar x).
\]
Writing $x^+=J_{\eta}(a)$, the resulting triple satisfies
$\bar x\in\X$, $\bar u=F(\bar x)$,
$\bar n\in N_{\X}(\bar x)$, and
\[
\norm{a-\bar x-\eta(\bar u+\bar n)}
\le
\eta L(1+\eta L)\norm{\tilde x-x^+}.
\]
Consequently, combining \Cref{cor:lipschitz-bridge}
with one projection yields a $\delta$-approximate
constrained resolvent selection in time
\[
\mathrm{poly}\bigl(
d,\log(L+\eta^{-1}),\log(1/\delta)
\bigr)+T_{\mathrm{proj}}(d).
\]

For single-valued Lipschitz Minty VIs, this provides
the oracle access required by \Cref{cor:reduction}.
Moreover, the strict GEVI certificates of
\Cref{thm:approx-disjunctive-appendix} reduce to
strict EVI certificates in this setting.

If the instance also satisfies the constrained
weak-Minty condition of
\citet{anagnostides2025polynomial} with
\[
\kappa\le C\frac{L}{(B+4RL)^2},
\qquad C<1,
\]
where $B:=\sup_{x\in\X}\norm{F(x)}$,
then \Cref{prop:weak-minty-robust} applies with
$\eta=1/(2L)$.
\end{corollary}

\begin{proof}
The projection optimality condition gives
$\bar x\in\X$ and $\bar n\in N_{\X}(\bar x)$,
while $\bar u=F(\bar x)$.
By \Cref{prop:single-valued-bridge}, the exact
resolvent point satisfies
\[
x^+=P_{\X}\bigl(a-\eta F(x^+)\bigr).
\]
The nonexpansiveness of Euclidean projection
and the Lipschitz continuity of $F$ imply
\[
\norm{\bar x-x^+}
\le
\eta\norm{F(\tilde x)-F(x^+)}
\le
\eta L\norm{\tilde x-x^+}.
\]
By the definitions of $\bar u$ and $\bar n$,
\[
a-\bar x-\eta(\bar u+\bar n)
=
\eta\bigl(F(\tilde x)-F(\bar x)\bigr).
\]
Hence,
\[
\begin{aligned}
\norm{a-\bar x-\eta(\bar u+\bar n)}
&\le \eta L\norm{\tilde x-\bar x}\\
&\le \eta L(1+\eta L)\norm{\tilde x-x^+}.
\end{aligned}
\]
Therefore, a point approximation satisfying
\[
\norm{\tilde x-x^+}
\le \frac{\delta}{\eta L(1+\eta L)}
\]
produces a $\delta$-approximate constrained
resolvent triple after one projection.
The runtime follows from
\Cref{cor:lipschitz-bridge}.

The Minty-SVI guarantee follows from
\Cref{cor:reduction}. Since $A(x)=\{F(x)\}$,
the strict GEVI alternative of
\Cref{thm:approx-disjunctive-appendix}
coincides with the strict EVI alternative.

For the weak-Minty claim, using $B\ge 0$
and $\eta=1/(2L)$ gives
\[
\kappa
\le \frac{CL}{(B+4RL)^2}
\le \frac{C}{16LR^2}
< \frac{1}{8LR^2}
= \frac{\eta}{4R^2}.
\]
Thus the admissibility condition of
\Cref{prop:weak-minty-robust} is satisfied.
\end{proof}

The two-player smooth concave-game construction in
\citet[Section~5]{anagnostides2025polynomial}
also leads to a single-valued Lipschitz VI on a compact
truncated domain. When Euclidean projection onto this
domain is efficient, the same argument provides the
required graph-residual resolvent oracle.
Furthermore, \citet[Lemma~5.3]{anagnostides2025polynomial}
show how sufficiently strict EVI certificates for the
transformed VI yield strict coarse correlated equilibria.

\section{Why the Direct Converse Fails Under Minty}
\label{app:false-converse}

The forward reduction established in \Cref{cor:reduction}
holds under the Minty condition alone and can be summarized as
\[
\text{efficient approximate resolvent}
\;\Longrightarrow\;
\text{polylog-accuracy Minty-SVI}.
\]
The reverse reduction requires additional structure.
In particular, applying a Minty-SVI solver to recover
$J_\eta(a)$ requires solving the anchored problem associated
with
\[
A_{a,\eta}(x):=A(x)+\frac{1}{\eta}(x-a).
\]
Under the Minty condition alone, this regularization
may have an empty Minty set even when $J_\eta(a)$ is
nonempty. The following proposition gives an explicit
counterexample.

\begin{proposition}
There exist a compact convex set $\X\subset\R$,
a single-valued affine operator $A:\X\to\R$,
and parameters $\eta>0$ and $a\in\R$ such that
\begin{enumerate}
    \item $A$ has a nonempty Minty set on $\X$;
    \item the regularized operator
    $A_{a,\eta}(x):=A(x)+\eta^{-1}(x-a)$
    has an empty Minty set on $\X$;
    \item the constrained resolvent set
    $J_\eta(a)$ is nonempty.
\end{enumerate}
\end{proposition}

\begin{proof}
Consider
\[
\X=[-1,1],\qquad
A(x)=-2x-3,\qquad
\eta=1,\qquad a=-3.
\]

Since $A(y)<0$ for every $y\in\X$, the point
$x^\star=1$ satisfies
\[
A(y)(y-1)\ge 0
\qquad\text{for all }y\in\X.
\]
Conversely, for any $x<1$, choosing $y=1$ gives
\[
A(1)(1-x)=-5(1-x)<0.
\]
Hence the Minty set of $A$ is $K=\{1\}$.

For the chosen anchor and scale, the regularized
operator becomes
\[
A_{a,1}(x)=A(x)+(x-a)=-x.
\]
If $x\in\X$ satisfied the Minty condition for
$A_{a,1}$, evaluating the Minty inequality at
$y=1$ and $y=-1$ would require
\[
(-1)(1-x)\ge 0
\quad\Longrightarrow\quad x\ge 1,
\]
and
\[
(1)(-1-x)\ge 0
\quad\Longrightarrow\quad x\le -1.
\]
These inequalities are incompatible, so the
Minty set of $A_{a,1}$ is empty.

Nevertheless, the constrained resolvent is
nonempty. Indeed, the relation
\[
a=x+A(x)+n,\qquad n\in N_{\X}(x),
\]
reduces to
\[
-3=x-2x-3+n,
\qquad\text{or equivalently}\qquad n=x.
\]
On $\X=[-1,1]$, this condition holds precisely
at $x\in\{-1,0,1\}$. Therefore,
\[
J_1(-3)=\{-1,0,1\}\neq\emptyset.
\]
\end{proof}

The example shows that Minty feasibility of $A$
does not guarantee Minty feasibility of its
anchored regularization. Consequently, a
Minty-SVI solver cannot be applied directly to
the regularized subproblem without additional
assumptions.

\section{Linear Convergence under a Yosida Error Bound}
\label{app:linear-rate}

Our ellipsoid method achieves polylogarithmic oracle complexity under the Minty condition alone, without requiring additional regularity. For monotone inclusions satisfying a Yosida error bound, however, the simpler fixed-scale resolvent iteration already converges linearly, making the cutting-plane procedure unnecessary.

\begin{definition}[Yosida error bound]
A monotone inclusion satisfies a \emph{Yosida error bound}
at scale $\eta$ with parameter $\kappa\in(0,1]$ if
\[
\eta\norm{g_{\eta}(a)}
\ge \kappa\,\dist(a,\Minty)
\qquad\text{for all }a\in\X.
\]
\end{definition}

\begin{theorem}
\label{thm:linear-rate}
Assume $A$ is monotone and satisfies the Yosida error
bound at scale $\eta$ with parameter $\kappa\in(0,1]$.
Starting from any $a_0\in\X$, consider the exact
resolvent iteration
\[
a_{t+1}:=J_{\eta}(a_t).
\]
Then
\[
\dist(a_{t+1},\Minty)^2
\le
(1-\kappa^2)\dist(a_t,\Minty)^2
\qquad\text{for all }t\ge 0.
\]
Consequently, an $\eps$-SVI can be computed in
\[
\mathcal{O}\!\left(
1+\kappa^{-2}
\log\left(1+\frac{R+\rho_{\eta}(\eps,R)}
{\rho_{\eta}(\eps,R)}\right)
\right)
\]
exact resolvent evaluations. In the regime
$\eps\le R^2/\eta$, this simplifies to
\[
\mathcal{O}\!\left(
\kappa^{-2}
\log\left(1+\frac{R^2}{\eta\eps}\right)
\right).
\]
For monotone $A$, the returned point also satisfies
the $\eps$-Minty inequalities.
\end{theorem}

\begin{proof}
Let $x_t^\star$ be a nearest Minty point to $a_t$.
By the resolvent identity and
\Cref{lem:exact-separation},
\begin{align*}
\dist(a_{t+1},\Minty)^2
&\le
\norm{a_t-x_t^\star-\eta g_{\eta}(a_t)}^2\\
&=
\dist(a_t,\Minty)^2
-2\eta\ip{g_{\eta}(a_t)}{a_t-x_t^\star}
+\eta^2\norm{g_{\eta}(a_t)}^2\\
&\le
\dist(a_t,\Minty)^2
-\eta^2\norm{g_{\eta}(a_t)}^2\\
&\le
(1-\kappa^2)\dist(a_t,\Minty)^2,
\end{align*}
where the last inequality follows from the
Yosida error bound.

The same separation inequality and
Cauchy--Schwarz imply
\[
\eta\norm{g_{\eta}(a_t)}
\le \dist(a_t,\Minty).
\]
Thus, if
$\dist(a_t,\Minty)\le\rho_{\eta}(\eps,R)$,
then
\[
s(a_t)=\eta\norm{g_{\eta}(a_t)}
\le\rho_{\eta}(\eps,R).
\]
By \Cref{lem:exact-upper},
\[
\Gap_{\eta}(a_t)
\le
\frac{s(a_t)^2+2Rs(a_t)}{\eta}
\le\eps.
\]
Hence \Cref{lem:gap-cert} certifies that
$a_{t+1}=J_{\eta}(a_t)$ is an $\eps$-SVI.

Since $\dist(a_0,\Minty)\le 2R$, the contraction
established above yields
\[
\dist(a_t,\Minty)
\le 2R\exp(-\kappa^2t/2).
\]
The stated iteration bounds follow by requiring
this distance to be at most
$\rho_{\eta}(\eps,R)$ and using
$\rho_{\eta}(\eps,R)=\Theta(\eta\eps/R)$
when $\eps\le R^2/\eta$.

Finally, let $u_{t+1}\in A(a_{t+1})$ be the
returned SVI witness. By monotonicity, for every
$y\in\X$ and $v\in A(y)$,
\[
\ip{v}{y-a_{t+1}}
\ge
\ip{u_{t+1}}{y-a_{t+1}}
\ge -\eps,
\]
which gives the $\eps$-Minty guarantee.
\end{proof}

Metric subregularity of the constrained operator
provides a sufficient condition for the Yosida
error bound.

\begin{proposition}
\label{prop:metric-subreg}
Assume $A$ is monotone and let
\[
T:=A+N_{\X},
\qquad
Z:=\set{x\in\X:0\in T(x)}\neq\emptyset.
\]
Suppose $T$ is globally metrically subregular
for $0$ on $\X$ with modulus $\lambda\ge 0$,
meaning that
\[
\dist(x,Z)
\le\lambda\,\dist(0,T(x))
\qquad\text{for all }x\in\X.
\]
Then, at any scale $\eta>0$ with exact
constrained-resolvent access, the Yosida error
bound holds with
\[
\kappa=\frac{\eta}{\eta+\lambda}.
\]
\end{proposition}

\begin{proof}
Fix $a\in\X$ and let
\[
x^+:=J_{\eta}(a),
\qquad
g_{\eta}(a)=\frac{a-x^+}{\eta}.
\]
Since $g_{\eta}(a)\in T(x^+)$, metric
subregularity gives
\[
\dist(x^+,Z)
\le \lambda\,\dist(0,T(x^+))
\le \lambda\norm{g_{\eta}(a)}.
\]
By the triangle inequality,
\[
\dist(a,Z)
&\le \norm{a-x^+}+\dist(x^+,Z)\\
&\le (\eta+\lambda)\norm{g_{\eta}(a)}.
\]
Since $A$ is monotone, $Z\subseteq\Minty$.
Consequently,
\[
\dist(a,\Minty)
\le\dist(a,Z)
\le(\eta+\lambda)\norm{g_{\eta}(a)}.
\]
Rearranging yields
\[
\eta\norm{g_{\eta}(a)}
\ge
\frac{\eta}{\eta+\lambda}\dist(a,\Minty),
\]
as claimed.
\end{proof}

\end{document}